\documentclass[11pt]{article}

\usepackage[final]{acl}

\usepackage{times}
\usepackage{latexsym}

\usepackage[T1]{fontenc}

\usepackage[utf8]{inputenc}

\usepackage{microtype}

\usepackage{inconsolata}

\usepackage{graphicx}

\title{Unveiling the Landscape of Clinical Depression Assessment:\\From Behavioral Signatures to Psychiatric Reasoning}

\author{
	\textbf{Zhuang Chen\textsuperscript{1}},
	\textbf{Guanqun Bi\textsuperscript{2}},
	\textbf{Wen Zhang\textsuperscript{3}},
	\textbf{Jiawei Hu\textsuperscript{4}},
	\\
	\textbf{Aoyun Wang\textsuperscript{1}},
	\textbf{Xiyao Xiao\textsuperscript{5}},
	\textbf{Kun Feng\textsuperscript{6}},
	\textbf{Minlie Huang\textsuperscript{2}}
	\\
	\\
	\textsuperscript{1}School of Computer Science and Engineering, Central South University\\
	\textsuperscript{2}CoAI Group, DCST, IAI, BNRIST, Tsinghua University\quad
	\textsuperscript{3}University of International Relations \\
	\textsuperscript{4}Central China Normal University \quad
	\textsuperscript{5}Lingxin AI \quad
	\textsuperscript{6}Yuquan Hospital, Tsinghua University\\
	{zhchen18@foxmail.com \quad aihuang@tsinghua.edu.cn}}

\usepackage{booktabs}
\usepackage{multirow}
\usepackage{graphicx}
\usepackage{subcaption}
\usepackage{xcolor}
\usepackage{colortbl}
\usepackage{enumitem}
\usepackage[most]{tcolorbox}
\usepackage{adjustbox}
\usepackage{tikz}

\begin{document}
\maketitle

\begin{abstract}

	Depression is a widespread mental disorder that affects millions worldwide. While automated depression assessment shows promise, most studies rely on limited or non-clinically validated data, and often prioritize complex model design over real-world effectiveness. 
	In this paper, we aim to unveil the landscape of clinical depression assessment. We introduce {C-MIND}, a \textit{clinical neuropsychiatric multimodal diagnosis} dataset collected over two years from real hospital visits. Each participant completes three structured psychiatric tasks and receives a final diagnosis from expert clinicians, with informative audio, video, transcript, and functional near-infrared spectroscopy (fNIRS) signals recorded.
	Using C-MIND, we first analyze \textit{behavioral signatures} relevant to diagnosis. We train a range of classical models to quantify how different tasks and modalities contribute to diagnostic performance, and dissect the effectiveness of their combinations. 
	We then explore whether LLMs can perform \textit{psychiatric reasoning} like clinicians and identify their clear limitations in realistic clinical settings. In response, we propose to guide the reasoning process with clinical expertise and consistently improves LLM diagnostic performance by up to 10\% in Macro-F1 score.
	We aim to build an infrastructure for clinical depression assessment from both data and algorithmic perspectives, enabling C-MIND to facilitate grounded and reliable research for mental healthcare.

\end{abstract}

\section{Introduction}

Depression is a widespread and serious mental disorder that places a heavy burden on individuals and public health systems worldwide. While automated assessment shows promise for offering objective and scalable support, its real-world clinical utility remains limited due to a lack of clinically grounded data \cite{cummins2015speech, sarsam2024multimodal}. Most widely used datasets rely on self-reported questionnaires rather than diagnoses made by trained clinicians \cite{gratch2014distress, tadesse2019detection}. Even the few pioneering studies that include clinical diagnoses often suffer from small sample sizes ($<$ 30 patients) and limited behavioral tasks or modalities \cite{cai2022multi, zou2022semi}. These constraints lead many studies to focus on sophisticated model design in controlled settings instead of addressing the full complexity of real clinical data. As a result, a clear picture of what effective automated clinical depression assessment entails has yet to emerge \cite{sarsam2024multimodal}.

\begin{figure}[t]
	\centering
	\includegraphics[width=0.47\textwidth]{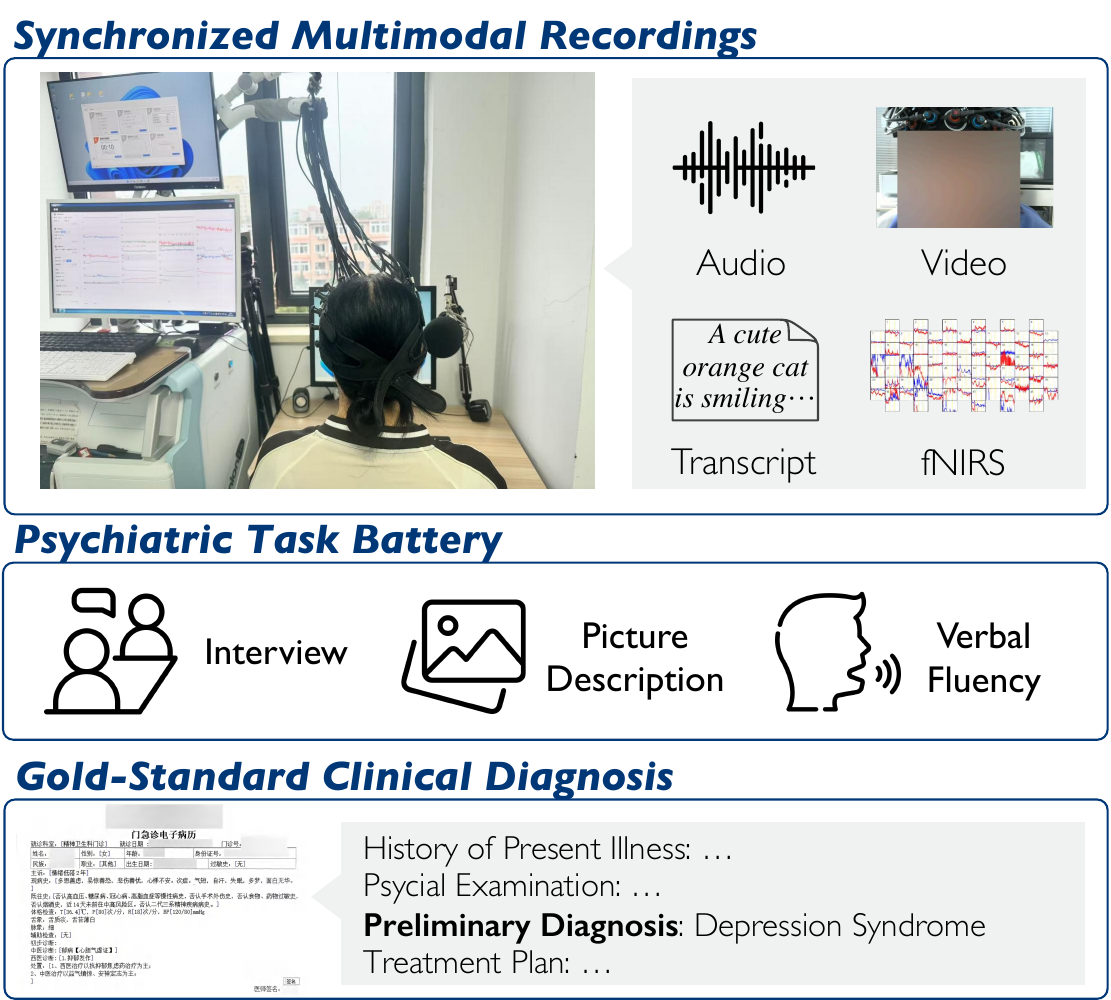}
	\caption{C-MIND integrates multimodal recordings from psychiatric tasks with clinical diagnosis.}
	\label{fig-example}
	\vspace{-4mm}
\end{figure}

\begin{figure*}[t]
	\centering
	\includegraphics[width=1\textwidth]{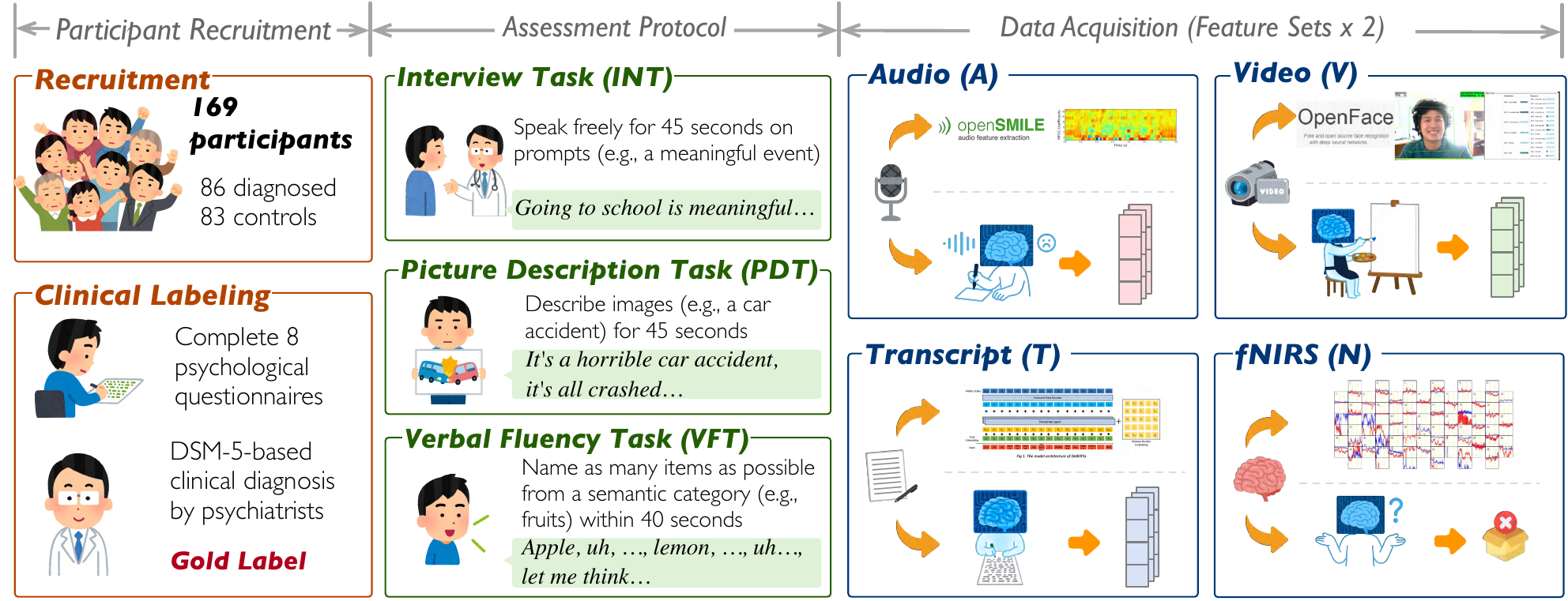}
	\caption{C-MIND collection pipeline, outlining participant recruitment, assessment protocol, and data acquisition. }
	\label{fig-collection}
	\vspace{-3mm}
\end{figure*}

In this paper, we aim to unveil this landscape through a three-pronged investigation: 1) establishing a new, clinically grounded data foundation, 2) analyzing the core behavioral signatures, and 3) advancing clinically guided psychiatric reasoning for diagnosis. First, we introduce \textbf{C-MIND}: the \underline{C}linical \underline{M}ult\underline{i}modal \underline{N}europsychiatric \underline{D}iagnosis dataset. Over a two-year period, we build this dataset from a real hospital setting. It comprises 169 participants who each complete three distinct psychiatric tasks, including Interview \cite{gratch2014distress}, Picture Description \cite{ramponi2010picture}, and Verbal Fluency \cite{fossati2003fluency}. We capture four synchronized modalities (Audio, Video, Transcript, and fNIRS \cite{cui2010fnirs}) for each session. Crucially, every participant receives a face-to-face diagnostic interview with senior psychiatrists, whose final clinical diagnosis, made according to DSM-5 \cite{APA2013} criteria, serves as the gold-standard ground truth. The dataset is further enriched with detailed medical records and a battery of eight psychometric questionnaires. C-MIND’s scale, clinical grounding, and richness in tasks and modalities far exceed previously available resources.

Leveraging C-MIND, we conduct an in-depth analysis of \textbf{behavioral signatures}, defined as observable patterns in speech, facial expression, and neural activity indicative of depressive states. We train a range of modeling backbones to systematically quantify the diagnostic value of different tasks and modalities, revealing that audio and video are the most informative, while the picture description task best elicits depressive markers. Fusing modalities (e.g., Audio+Video) or tasks (e.g., Interview+Picture Description) further enhances performance and robustness, providing clear empirical guidance for designing future assessment systems.

Beyond analyzing predictive signals, we explore whether Large Language Models (LLMs) perform \textbf{psychiatric reasoning}. We evaluate seven top-tier text and multimodal LLMs and find clear limitations in their ability to handle real-world clinical data. In response, we propose a novel method that guides the LLM’s reasoning process using structured clinical expertise. This approach significantly boosts diagnostic performance by up to 10\% in Macro-F1 score, demonstrating a promising direction for developing clinically informed computational models.

Our main contributions can be summarized as follows:
\begin{itemize}[leftmargin=1em, topsep=0pt, itemsep=0pt, parsep=0pt]
	\item We introduce C-MIND, a clinically validated depression diagnosis dataset with rich tasks and modalities.
	\item We provide a comprehensive analysis of behavioral signatures, offering clear, data-driven insights into the discriminative power of different tasks and modalities.
	\item We demonstrate the limitations of LLMs in clinical assessment and propose a novel psychiatric reasoning mechanism that significantly boosts performance.
\end{itemize}
We believe this work builds a critical infrastructure for the field and provides a blueprint for developing computational systems that are not only effective, but also clinically grounded and trustworthy.

\section{C-MIND Collection}
\label{sec:dataset}

We present \textbf{C-MIND}, a Clinical Multimodal Neuropsychiatric Diagnosis dataset designed for depression assessment. To ensure ecological validity, data quality, and clinical reliability, we follow a comprehensive collection protocol. Below, we describe participant recruitment, assessment procedures, and data acquisition in detail.

\subsection{Participant Recruitment}

We recruited participants from December 2022 to April 2025 at the psychiatric department of a university-affiliated hospital. Recruitment was conducted through internal announcements. Volunteers who met the inclusion criteria and were evaluated by a chief psychiatrist together with an associate chief psychiatrist according to DSM-5~\cite{APA2013} were invited to participate after providing written informed consent.

\begin{table}[t]
	\vspace{0em}
	\centering	
	\renewcommand{\arraystretch}{1.2}
	\setlength{\tabcolsep}{1mm}
	\resizebox{0.475\textwidth}{!}{%
		\begin{tabular}{lccc}
			\toprule
			\textbf{Statistic} & \textbf{Depression} & \textbf{Control} & \textbf{Total} \\
			\midrule
			\textbf{Subject} & 86 & 83 & 169 \\
			\textbf{Gender (M/F \%)} & 36.05/63.95 & 44.58/55.42 & 40.34/59.66 \\
			\textbf{Age (Mean$\pm$SD)} & 33.49 $\pm$16.47 & 32.47$\pm$10.90 & 32.99$\pm$13.98 \\
			\textbf{Duration (s)} & 171.84 & 125.01 & 152.16 \\
			\textbf{Word Count} & 392 & 519 & 445 \\
			\bottomrule
		\end{tabular}
	}
	\caption{Detailed statistics of C-MIND.} 
\label{tab:demographics}
\vspace{-1em}

\end{table}

\begin{table*}[!t]
	\centering
	\setlength{\tabcolsep}{1mm}
	\renewcommand{\arraystretch}{1.15} 
	\resizebox{\textwidth}{!}{%
		\begin{tabular}{clcccccc}
			\toprule
			\textbf{Availability} & \textbf{Dataset} & \textbf{Language} & \textbf{Subj. (MDD/HC)} & \textbf{Tasks} & \textbf{Modalities} & \textbf{Ground Truth Labels} \\
			\midrule
			\multirow{4}{*}{No} 
			& Oizys~\cite{lin2022deep}         & Chinese & 103 (56/47)   & READ             & A         & C.D., HAMD-17           \\
			& \citet{guo2021deep}              & Chinese & 208 (104/104) & INT, READ, PDT    & A, V      & PHQ-9, BDI            \\
			& \citet{liu2021improved}          & Chinese & 50 (25/25)    & INT              & A         & BDI                   \\
			& DEPAC~\cite{tasnim2022depac}     & English & 552 (134/418) & VFT, PDT, READ     & A         & PHQ-9, GAD-7          \\
			\midrule
			\multirow{2}{*}{\shortstack{Yes \\ (\textit{w/o C.D.})}}
			& DAIC-WOZ~\cite{gratch2014distress}  & English & 142 (42/100)  & INT              & A, V, T   & PHQ-8                 \\
			& EATD (Shen et al. 2022)       & Chinese & 162 (30/132)  & INT              & A, T      & SDS                   \\
			\midrule
			\multirow{3}{*}{\shortstack{Yes \\ (\textit{with C.D.})}}
			& MODMA~\cite{cai2022multi}           & Chinese & 53 (24/29)    & INT, READ, PDT    & A, EEG    & C.D., PHQ-9             \\
			& CMDC~\cite{zou2022semi}             & Chinese & 78 (26/52)    & INT              & A, V, T   & C.D.$^-$, HAMD-17, PHQ-9\\
			& \textbf{C-MIND (Ours)}              & Chinese & \textbf{169 (86/83)}   & \textbf{INT, PDT, VFT}      & \textbf{A, V, T, fNIRS} & \textbf{C.D.$^+$, 8 Questionnaires}    \\
			\bottomrule
		\end{tabular}
	}
	\caption{Comparison of depression datasets. C.D.=Clinical Diagnosis, INT=Interview, PDT=Picture Description, VFT=Verbal Fluency, READ=Reading,  A=Audio, V=Video, T=Transcript, fNIRS=functional near-infrared spectroscopy. ``C.D.$^-$'' means the control group is not confirmed by clinical diagnosis. ``C.D.$^+$'' means our C-MIND further provides detailed medical records.}
	\label{tab:dataset_comparison}
\end{table*}

All procedures receive full approval from the university’s Institutional Review Board (IRB), and strict measures are in place to protect participant confidentiality. The final cohort consists of 169 participants, including 86 individuals diagnosed with Major Depressive Disorder (MDD) and 83 healthy controls (HC). Table~\ref{tab:demographics} presents detailed statistics of the C-MIND cohort, including group size, gender distribution, age, average speech duration, and word count.
For future public release, we will follow IRB-approved protocols to ensure responsible data sharing, including strict de-identification and a request-based access process.

\subsection{Assessment Protocol}

The assessment protocol for each participant includes two main parts: a formal face-to-face clinical diagnosis (used to obtain clinically validated labels), and a series of psychiatric tasks (used to collect rich multimodal behavioral signatures). Due to space constraints, we provide detailed experimental materials, guidelines, and procedures in the Technical Appendix.

\subsubsection{Clinical Diagnosis}

Participants are interviewed by a clinical team comprising a chief and an associate chief psychiatrist, each with over ten years of experience. The team conducts a face-to-face diagnostic interview and makes a high-confidence diagnosis based on DSM-5 criteria. A detailed medical record is maintained for each participant. 
Participants also complete a battery of eight psychometric questionnaires, including: \textit{HAMD}~\cite{Hamilton1960}, \textit{HAMA}~\cite{Hamilton1959}, \textit{SDS}~\cite{Zung1965}, \textit{SAS}~\cite{Zung1971}, \textit{PSQI}~\cite{Buysse1989}, \textit{16PF}~\cite{Cattell1970}, \textit{SCL-90}~\cite{Derogatis1977}, and \textit{HCL-32}~\cite{Angst2005}. These instruments assess depressive symptoms, anxiety, sleep quality, and personality traits. In this study, we use only the clinical diagnosis as the ground-truth label; questionnaire data are reserved for future work.

\subsubsection{Psychiatric Tasks}

All tasks take place in a quiet, controlled laboratory with ambient noise below 60dB. Participants sit in front of a monitor displaying instructions and are asked to remain seated, minimize movement, and maintain a fixed distance from the microphone (approx. 20 cm). We design three structured tasks that elicit cognitive and emotional markers of depression:

\begin{itemize}[leftmargin=1em, topsep=0pt, itemsep=0pt, parsep=0pt]
	\item \textbf{Interview Task (INT):} Participants speak for 45 seconds in response to autobiographical prompts. This task elicits emotional expression and narrative patterns indicative of depression (e.g., negative sentiment, frequent first-person use)~\cite{Rinaldi2020}. Prompts include: ``something that made you angry,'' ``a meaningful event,'' and ``your favorite food.''
	
	\item \textbf{Picture Description Task (PDT):} Participants describe a given image for 45 seconds. This task captures visual interpretation and emotional valence. Depressed individuals may show negative bias or limited detail~\cite{Ramponi2010}. Images include: ``a cat,'' ``a car accident,'' and ``a spaceship.''
	
	\item \textbf{Verbal Fluency Task (VFT):} Participants list items in a semantic category (e.g., ``fruits'') within 40 seconds. This evaluates semantic memory and executive function, which are often impaired in depression~\cite{Akiyama2018,fossati2003fluency}. Categories include: ``four-legged animals,'' ``fruits,'' ``cities,'' and ``vegetables.''
\end{itemize}

In total, each participant completes ten tests across the three tasks. We record four synchronized modalities during the psychiatric tasks. We use a studio microphone to capture high-quality audio (44.1kHz, 24-bit, WAV). A camera records video (640x480, 30fps), and a functional near-infrared spectroscopy (fNIRS) device measures blood oxygenation changes in the prefrontal cortex. Each of the 10 tasks is recorded separately in audio. Using timestamps, we segment the continuous video and fNIRS recordings to align all modalities. Transcripts are generated using a commercial speech recognition system and are manually proofread by human annotators to ensure accuracy.

\subsection{Comparison with Existing Datasets}

To situate C-MIND within the current research landscape, we compare it with other depression datasets collected in controlled environments, while excluding those based on social media data~\cite{yoon2022dvlog}.

As summarized in Table~\ref{tab:dataset_comparison}, many existing datasets (e.g., DAIC-WOZ~\cite{gratch2014distress}) rely on self-report instruments like PHQ-8 and lack clinical validation. While MODMA~\cite{cai2022multi} and CMDC~\cite{zou2022semi} include clinical diagnoses, they have smaller sample sizes (53/78 participants with only 24/26 patients, respectively), limited tasks, and fewer modalities.

In contrast, C-MIND goes far beyond existing resources in every aspect: it offers a larger and balanced sample size, more diverse psychiatric tasks (INT, PDT, VFT), richer modalities (Audio, Video, Text, fNIRS), expert-verified clinical diagnoses, and comprehensive psychometric data—making it a uniquely comprehensive and clinically grounded benchmark for depression research.

\section{Methodology}

\begin{figure*}[t]
	\centering
	\includegraphics[width=1.0\textwidth]{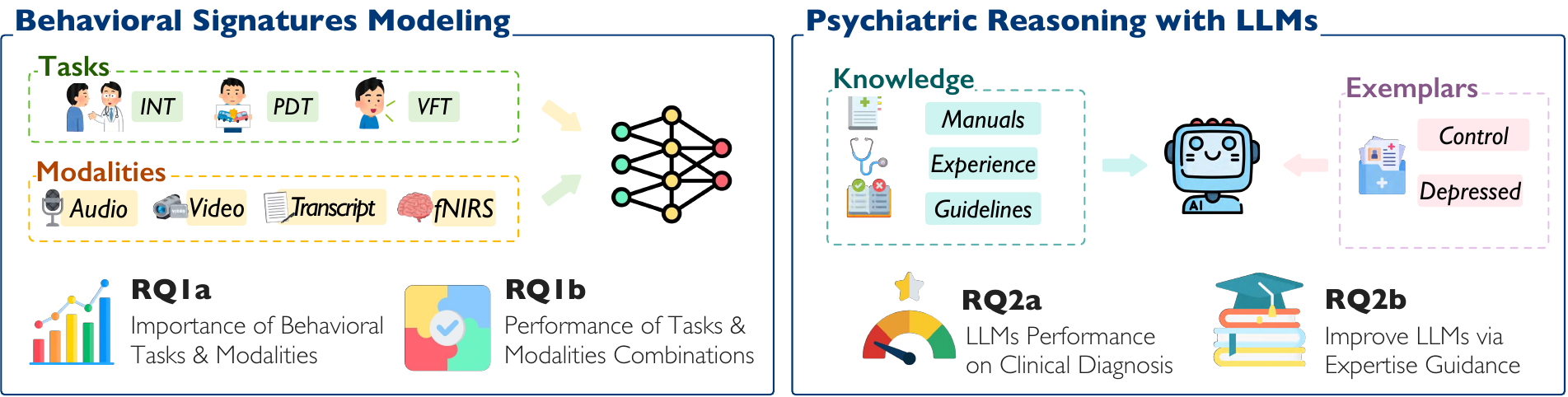}
	\caption{Overview of the two-part research methodology, addressing: 1) the diagnostic value of behavioral signatures (RQ1) and 2) the performance and enhancement of psychiatric reasoning in LLMs (RQ2).}
	\label{fig:method}
\end{figure*}

As shown in Figure \ref{fig:method}, we design a two-part methodological framework to uncover the mechanisms of clinical depression assessment: 1) modeling behavioral signatures across tasks and modalities, and 2) simulating psychiatric reasoning through guided LLMs. Due to space limitations, we present only core formulations here; details of models and prompts can be found in Technical Appendix.

\subsection{Behavioral Signature Modeling}
We define behavioral signatures as the informational cues derived from different psychiatric tasks and data modalities. 
To quantify the diagnostic relevance of different behavioral cues, we follow a structured modeling pipeline. First, we extract feature representations from four modalities (audio, video, text, fNIRS) for each psychiatric task. Then, we train learning models to predict depression status based on feature sets. By evaluating each task-modality combination and their fusions, we aim to uncover which behavioral signatures most robustly reflect clinical diagnoses. The entire process is designed to simulate and analyze how observable behaviors align with psychiatric assessment.

\paragraph{Feature Representations}
We extract two feature sets from four synchronized modalities: audio, video, transcript, and fNIRS. \textbf{1) Classical Feature Set.} We apply standard feature extraction pipelines. Audio signals are encoded using OpenSmile’s eGeMAPS (88 dimensions), while video-based facial behavior is represented using OpenFace (4,963 dimensions), including action units, gaze, and head pose. Textual transcripts are embedded using DeBERTa. For fNIRS, statistical features are computed from 45 optical channels, resulting in a 630-dimensional vector. \textbf{2) Foundation Model Feature Set.} We also extract semantic-level embeddings using pretrained foundation models: Qwen2-Audio-7B \cite{chu2023qwen} for audio, Qwen2.5-VL-72B for video \cite{bai2025qwen2}, and Qwen3-235B-A22B for transcript \cite{yang2025qwen3}. Each modality is segmented by task, encoded through the model’s final hidden layer, and globally max-pooled over time to yield fixed-length vectors (4096 dimensions for audio and transcript, 8192 for video). fNIRS remains represented by classical statistics due to the absence of public pretrained encoders.

\paragraph{Task \& Modality Modeling}
We denote the task set as $\mathcal{T} = \{\text{INT}, \text{PDT}, \text{VFT}\}$ and the modality set as $\mathcal{M} = \{\text{Audio}, \text{Video}, \text{Transcript}, \text{fNIRS}\}$. For each subject $i$, let $X^{(i)}_{m,t}$ represent the features from modality $m$ and task $t$. We train a classifier to map them to the clinical label:
\[
f_{\theta}: X^{(i)}_{m,t} \rightarrow y^{(i)}, \quad y^{(i)} \in \{0, 1\}
\]
This allows us to quantify the diagnostic value of each task-modality pair. To further explore whether aggregating behavioral evidence enhances performance, we conduct two types of fusion experiments. In \textit{task fusion}, we fix the modality and concatenate features across all tasks: $X^{(i)}_{\text{fused}} = \text{Concat}(X^{(i)}_{m,t_1}, \dots, X^{(i)}_{m,t_k})$ for every $m \in \mathcal{M}$. This setup allows us to assess whether different task designs provide complementary cognitive and affective signals. In \textit{modality fusion}, we fix the task set and concatenate features across all modalities: $X^{(i)}_{\text{fused}} = \text{Concat}(X^{(i)}_{m_1,t}, \dots, X^{(i)}_{m_k,t})$ for every $t \in \mathcal{T}$, followed by aggregation across tasks. This setting evaluates how multimodal observations enhance detection when applied consistently across structured clinical tasks. In both settings, the fused representation is passed to the same predictive function $f_\theta$ for classification.

\subsection{Psychiatric Reasoning with LLMs}

We examine whether LLMs can emulate clinician-like diagnostic reasoning based solely on transcripts (for text LLMs) or multimodal signals (for MLLM) from three structured psychiatric tasks: interview, picture description, and verbal fluency, spanning a total of ten tests (T1–T10). Each subject’s signals are concatenated into a single prompt, and the LLM is tasked with predicting a binary diagnostic label.

We compare three reasoning strategies. \textbf{1) Direct Prediction} asks the model to infer the diagnosis directly from the input without any explanation. \textbf{2) Vanilla Reasoning} encourages the model to engage in free-form, step-by-step reasoning before making a prediction. \textbf{3) Psychiatric Reasoning} is our proposed method, which guides the model through a structured reasoning process grounded in clinical expertise. This approach reflects our key insight: effective psychiatric diagnosis depends not only on what is said, but also on how it is expressed under different task demands. To operationalize this, the prompt incorporates task definitions and expert-informed behavioral expectations, helping the LLM attend to symptom-relevant cues in a way that aligns with real clinical reasoning. The essential structure is summarized below.
\begin{tcolorbox}[colback=gray!10!white, colframe=gray!50!white, sharp corners=south, boxrule=0.4pt]
	\small
	\textit{\textbf{Interview tasks} probe emotional tone and autobiographical specificity... \textbf{Picture description tasks} assess imagination, semantic flow, and emotional valence... \textbf{Verbal fluency tasks} test lexical diversity and cognitive flexibility... \textbf{Depressed individuals} often express negative sentiment, lack detail, and repeat or simplify content... \textbf{Healthy individuals} tend to produce specific, emotionally rich, and well-organized language...}
\end{tcolorbox}
Let $\mathcal{P}^{(i)} = \{T_1^{(i)}, \dots, T_{10}^{(i)}\}$ represent the transcript prompt for subject $i$. The LLM performs a binary classification $g_\phi(\mathcal{P}^{(i)}) \rightarrow y^{(i)} \in \{0,1\}$. When clinical guidance $\mathcal{K}$ is included, the model performs $g_\phi(\mathcal{P}^{(i)}, \mathcal{K}) \rightarrow y^{(i)}$, using the embedded knowledge to attend to diagnostically meaningful patterns.

\section{Experiments \& Analysis}
\label{sec:experiments}

\begin{table*}[t]
	\setlength{\abovecaptionskip}{1mm}
	\setlength{\belowcaptionskip}{0mm}
	\setlength{\tabcolsep}{0.5mm}
	\small
	\centering
	\renewcommand{\arraystretch}{1.4}
	\resizebox{0.9999\textwidth}{!}{%
		\begin{tabular}{cc|cccccc|cccccc|cccccc}
			\toprule
			\multirow{2}{*}{\textbf{Modality}} & \multirow{2}{*}{\textbf{Feature Set}}
			& \multicolumn{6}{c|}{\textbf{Interview (INT)}}
			& \multicolumn{6}{c|}{\textbf{Picture Description (PDT)}}
			& \multicolumn{6}{c}{\textbf{Verbal Fluency (VFT)}} \\
			& & LSTM & CNN & MLP & k-NN & RF & SVM & LSTM & CNN & MLP & k-NN & RF & SVM & LSTM & CNN & MLP & k-NN & RF & SVM \\
			\midrule
			\multirow{2}{*}{\textbf{Audio (A)}}
			& OpenSmile
			& \cellcolor[HTML]{B8DAE9}72.15 & \cellcolor[HTML]{B8DAE9}84.25 & \cellcolor[HTML]{B8DAE9}82.31 & \cellcolor[HTML]{B8DAE9}85.18 & \cellcolor[HTML]{B8DAE9}91.17 & \cellcolor[HTML]{B8DAE9}91.11
			& \cellcolor[HTML]{6BAED6}69.49 & \cellcolor[HTML]{6BAED6}88.24 & \cellcolor[HTML]{6BAED6}81.21 & \cellcolor[HTML]{6BAED6}94.10 & \cellcolor[HTML]{6BAED6}88.24 & \cellcolor[HTML]{6BAED6}85.18
			& \cellcolor[HTML]{7FB8DD}70.90 & \cellcolor[HTML]{7FB8DD}84.28 & \cellcolor[HTML]{7FB8DD}79.39 & \cellcolor[HTML]{7FB8DD}76.14 & \cellcolor[HTML]{7FB8DD}88.24 & \cellcolor[HTML]{7FB8DD}82.29 \\
			& Qwen-Audio
			& \cellcolor[HTML]{B8DAE9}72.57 & \cellcolor[HTML]{B8DAE9}78.41 & \cellcolor[HTML]{B8DAE9}58.39 & \cellcolor[HTML]{B8DAE9}55.54 & \cellcolor[HTML]{B8DAE9}69.26 & \cellcolor[HTML]{B8DAE9}76.39
			& \cellcolor[HTML]{6BAED6}74.07 & \cellcolor[HTML]{6BAED6}71.54 & \cellcolor[HTML]{6BAED6}67.58 & \cellcolor[HTML]{6BAED6}69.64 & \cellcolor[HTML]{6BAED6}72.47 & \cellcolor[HTML]{6BAED6}79.39
			& \cellcolor[HTML]{7FB8DD}76.93 & \cellcolor[HTML]{7FB8DD}82.33 & \cellcolor[HTML]{7FB8DD}61.63 & \cellcolor[HTML]{7FB8DD}78.96 & \cellcolor[HTML]{7FB8DD}76.43 & \cellcolor[HTML]{7FB8DD}76.39 \\
			\midrule
			\multirow{2}{*}{\textbf{Video (V)}}
			& OpenFace
			& \cellcolor[HTML]{A8D1E4}70.41 & \cellcolor[HTML]{A8D1E4}71.47 & \cellcolor[HTML]{A8D1E4}69.00 & \cellcolor[HTML]{A8D1E4}72.94 & \cellcolor[HTML]{A8D1E4}72.40 & \cellcolor[HTML]{A8D1E4}73.32
			& \cellcolor[HTML]{9ECAE1}72.25 & \cellcolor[HTML]{9ECAE1}75.46 & \cellcolor[HTML]{9ECAE1}65.42 & \cellcolor[HTML]{9ECAE1}69.64 & \cellcolor[HTML]{9ECAE1}74.49 & \cellcolor[HTML]{9ECAE1}64.58
			& \cellcolor[HTML]{C6DBEF}74.64 & \cellcolor[HTML]{C6DBEF}74.49 & \cellcolor[HTML]{C6DBEF}69.54 & \cellcolor[HTML]{C6DBEF}69.64 & \cellcolor[HTML]{C6DBEF}75.43 & \cellcolor[HTML]{C6DBEF}64.71 \\
			& Qwen-VL
			& \cellcolor[HTML]{A8D1E4}79.97 & \cellcolor[HTML]{A8D1E4}83.31 & \cellcolor[HTML]{A8D1E4}76.92 & \cellcolor[HTML]{A8D1E4}85.28 & \cellcolor[HTML]{A8D1E4}84.28 & \cellcolor[HTML]{A8D1E4}85.28
			& \cellcolor[HTML]{9ECAE1}79.91 & \cellcolor[HTML]{9ECAE1}86.27 & \cellcolor[HTML]{9ECAE1}80.32 & \cellcolor[HTML]{9ECAE1}88.24 & \cellcolor[HTML]{9ECAE1}83.29 & \cellcolor[HTML]{9ECAE1}85.28
			& \cellcolor[HTML]{C6DBEF}78.73 & \cellcolor[HTML]{C6DBEF}84.25 & \cellcolor[HTML]{C6DBEF}74.26 & \cellcolor[HTML]{C6DBEF}67.39 & \cellcolor[HTML]{C6DBEF}81.28 & \cellcolor[HTML]{C6DBEF}85.28 \\
			\midrule
			\multirow{2}{*}{\textbf{Transcript (T)}}
			& DeBERTa
			& \cellcolor[HTML]{F7FBFF}60.93 & \cellcolor[HTML]{F7FBFF}64.33 & \cellcolor[HTML]{F7FBFF}51.92 & \cellcolor[HTML]{F7FBFF}46.32 & \cellcolor[HTML]{F7FBFF}58.02 & \cellcolor[HTML]{F7FBFF}52.28
			& \cellcolor[HTML]{E6F0F9}66.36 & \cellcolor[HTML]{E6F0F9}62.16 & \cellcolor[HTML]{E6F0F9}48.54 & \cellcolor[HTML]{E6F0F9}55.54 & \cellcolor[HTML]{E6F0F9}52.87 & \cellcolor[HTML]{E6F0F9}51.43
			& \cellcolor[HTML]{EFF5FB}57.83 & \cellcolor[HTML]{EFF5FB}56.54 & \cellcolor[HTML]{EFF5FB}56.37 & \cellcolor[HTML]{EFF5FB}62.64 & \cellcolor[HTML]{EFF5FB}56.44 & \cellcolor[HTML]{EFF5FB}55.84 \\
			& Qwen
			& \cellcolor[HTML]{F7FBFF}60.57 & \cellcolor[HTML]{F7FBFF}64.69 & \cellcolor[HTML]{F7FBFF}58.16 & \cellcolor[HTML]{F7FBFF}60.92 & \cellcolor[HTML]{F7FBFF}61.37 & \cellcolor[HTML]{F7FBFF}61.73
			& \cellcolor[HTML]{E6F0F9}67.69 & \cellcolor[HTML]{E6F0F9}68.38 & \cellcolor[HTML]{E6F0F9}56.81 & \cellcolor[HTML]{E6F0F9}58.88 & \cellcolor[HTML]{E6F0F9}67.58 & \cellcolor[HTML]{E6F0F9}67.39
			& \cellcolor[HTML]{EFF5FB}55.07 & \cellcolor[HTML]{EFF5FB}59.29 & \cellcolor[HTML]{EFF5FB}53.83 & \cellcolor[HTML]{EFF5FB}76.14 & \cellcolor[HTML]{EFF5FB}62.13 & \cellcolor[HTML]{EFF5FB}64.21 \\
			\midrule
			\textbf{fNIRS (N)} & Statistics
			& \cellcolor[HTML]{D5E4F5}62.54 & \cellcolor[HTML]{D5E4F5}71.27 & \cellcolor[HTML]{D5E4F5}59.61 & \cellcolor[HTML]{D5E4F5}43.33 & \cellcolor[HTML]{D5E4F5}73.49 & \cellcolor[HTML]{D5E4F5}61.46
			& \cellcolor[HTML]{F7FBFF}65.98 & \cellcolor[HTML]{F7FBFF}69.06 & \cellcolor[HTML]{F7FBFF}54.83 & \cellcolor[HTML]{F7FBFF}42.05 & \cellcolor[HTML]{F7FBFF}73.30 & \cellcolor[HTML]{F7FBFF}45.36
			& \cellcolor[HTML]{DEEBF7}62.55 & \cellcolor[HTML]{DEEBF7}64.05 & \cellcolor[HTML]{DEEBF7}65.57 & \cellcolor[HTML]{DEEBF7}50.18 & \cellcolor[HTML]{DEEBF7}75.43 & \cellcolor[HTML]{DEEBF7}46.88 \\
			\bottomrule
		\end{tabular}%
	}
	\caption{Performance (Macro-F1) of different models and feature sets, evaluated per task-modality combination. The color of each block corresponds to the average performance of the results within that block (a darker shade indicates a higher average).}
	\label{feature_performance}
	\vspace{-2mm}
\end{table*}

\begin{figure*}[t]
	\centering	
	\setlength{\abovecaptionskip}{0mm}
	\setlength{\belowcaptionskip}{0mm}
	\includegraphics[width=0.999\textwidth]{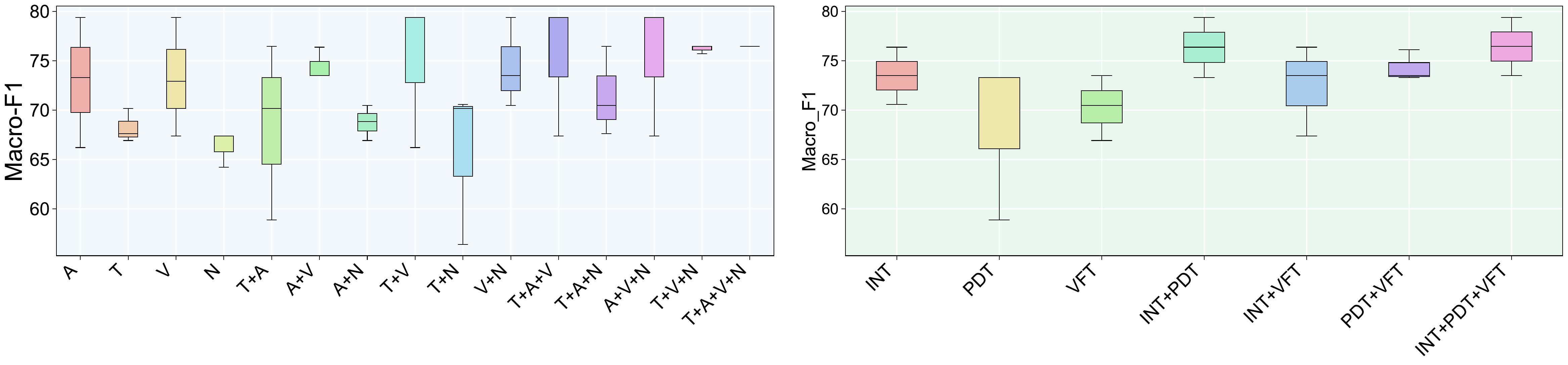}
	\caption{Performance (Macro-F1) of modality fusion (left) and task fusion (right). Combining signals generally improves the Macro-F1 score and reduces performance variance.}
	\label{fig:task_and_modal}
\end{figure*}

\begin{figure*}[t]
	\centering
	\includegraphics[width=1.0\textwidth]{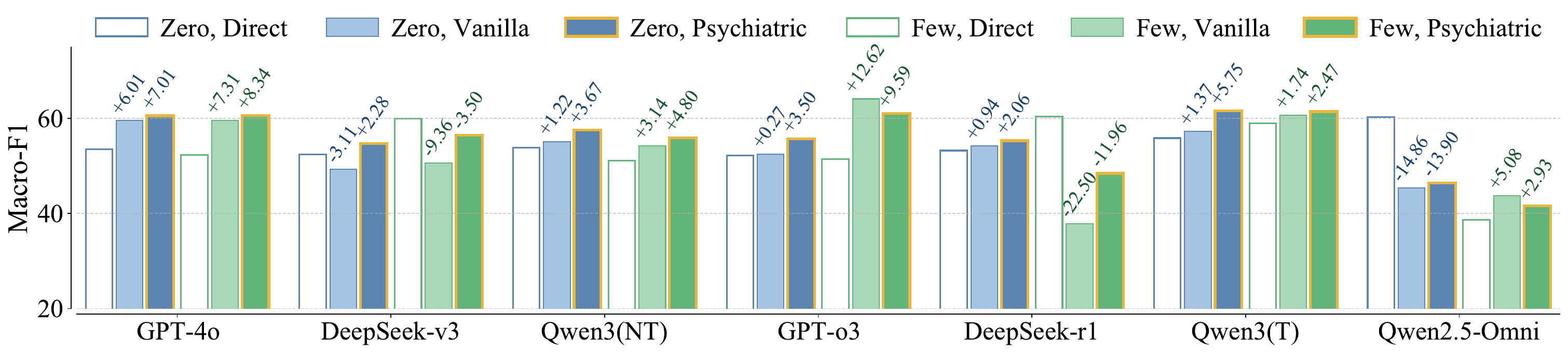}
	\caption{Performance comparison of LLMs using three reasoning strategies in zero/few-shot settings. Numbers above the bars indicate the performance change relative to the corresponding "Direct" baseline.}
	\label{fig:llm}
	\vspace{0mm}
\end{figure*}

\subsection{Experimental Settings}
We conduct all experiments on C-MIND. To ensure robust evaluation, we randomly split the dataset into training, validation, and test sets following a 6:2:2 ratio. We report Macro-F1 as the main evaluation metric. Full metrics, including Precision, Recall, and per-class F1 scores, are available in the Technical Appendix. All results are averaged over five independent runs with different random seeds. Our analysis aims to answer two key research questions (RQs):
\begin{itemize}[leftmargin=1em, topsep=2pt, itemsep=0pt, parsep=0pt]
	\item \textbf{RQ1:} What are the contributions of different behavioral tasks and modalities to depression assessment?
	\item \textbf{RQ2:} Can LLMs reason like clinical psychiatrists, and how can knowledge injection improve their performance?
\end{itemize}
To address RQ1, we benchmark a suite of classical learning backbones, including LSTM, CNN, MLP, k-NN, Random Forest (RF), and SVM. To address RQ2, we evaluate several leading LLMs. The text-based LLMs, including \texttt{GPT-4o}, \texttt{GPT-o3}, \texttt{DeepSeek-V3}, \texttt{DeepSeek-R1}, and \texttt{Qwen3-235B-A22B-T/NT} (thinking/non-thinking mode), use only the transcript as input. In contrast, the multimodal model \texttt{Qwen2.5-Omni} processes a combination of audio, video, and transcript. 
Due to space limitations, detailed model architectures, parameters, and versions are provided in the Technical Appendix.

\subsection{RQ1: The Power of Behavioral Signatures}

\paragraph{Tasks and Modalities} As shown in Table \ref{feature_performance}, we evaluate each task and modality using two feature sets. Our analysis reveals that Audio and Video are the most informative modalities, though their effectiveness is deeply intertwined with the psychiatric task being performed. Each task is designed to probe different cognitive and emotional facets, and their diagnostic power comes from how well these probes elicit observable, depression-related behavioral markers.

The Picture Description Task (PDT), for instance, excels in this regard, proving to be the most effective probe in our analysis. This is clinically intuitive as it assesses for emotional and attentional biases. Depressed individuals may exhibit a negative interpretation bias or provide less detailed descriptions, which is reflected not only in word choice (Transcript) but crucially in a flat vocal tone (Audio) and blunted affect (Video). This is evidenced by the top-performing model, which achieved a 94.10\% Macro-F1 score using audio features from the PDT. Similarly, the Verbal Fluency Task (VFT) also shows remarkable performance, particularly with audio features. VFT assesses executive functions and semantic memory, which are often impaired in depression. This cognitive deficit doesn't just manifest as a lower word count, but more saliently as acoustic patterns like longer pauses, frequent hesitation markers (e.g., "uh", "um"), and reduced prosodic variation. These are precisely the signals captured by audio analysis, explaining its success. The Interview task (INT) remains a robust baseline because its autobiographical prompts are effective at eliciting narratives laden with depressive markers like negative sentiment and overgeneralization, signals that are present across Audio, Video, and Transcript.

\paragraph{Fusion Improves Robustness} As illustrated in Figure \ref{fig:task_and_modal}, a clear and consistent finding is that fusing evidence from multiple sources enhances diagnostic performance. Combining modalities (e.g., Audio and Video) or integrating tasks (e.g., INT and PDT) consistently leads to higher Macro-F1 scores and, critically, more stable and reliable predictions by reducing variance. This underscores the value of a holistic assessment strategy, where a richer, multi-faceted view of a participant's behavior provides a more robust foundation for clinical inference than any single signal alone.

\subsection{RQ2: Psychiatric Reasoning with LLMs}

\paragraph{Reasoning Strategies}
We evaluate seven leading LLMs under three prompting strategies: \textit{Direct Prediction}, \textit{Vanilla Reasoning}, and our proposed \textit{Psychiatric Reasoning}, across both zero-shot and few-shot conditions (Figure~\ref{fig:llm}). Several consistent patterns emerge.

1) {Psychiatric Reasoning consistently improves zero-shot performance.} Across most non-thinking models (e.g., GPT-4o, GPT-o3, Qwen3(NT)), the structured psychiatric prompt yields stable gains (e.g., +7.01\% for GPT-4o, +3.67\% for Qwen3(NT)), outperforming both Direct and Vanilla strategies. 
2) {Few-shot performance gains vary, and can conflict with structured guidance.} For GPT-4o, Vanilla Reasoning under few-shot improves significantly (+7.31\%), but the gain from Psychiatric Reasoning (+8.34\%) suggests that explicit guidance remains beneficial. In contrast, DeepSeek-v3 and DeepSeek-r1 show degradation under few-shot reasoning, likely due to incompatibility between pretrained reasoning paths and injected prompts. 
3) {Models with internal reasoning protocols may conflict with external prompts.} DeepSeek-r1 exhibits a significant drop when reasoning is added, especially under few-shot settings (-22.5\% with Vanilla and -11.96\% with Psychiatric prompts), highlighting potential interference from overlaying external logic on built-in reasoning. 
4) {Multimodal input does not guarantee improved performance.} The multimodal model Qwen2.5-Omni consistently underperforms across all prompting settings, achieving just 46.36\% with Psychiatric Reasoning (zero-shot), which is worse than most transcript-only models. This suggests that general-purpose multimodal LLMs currently lack the fine-grained capability to utilize clinical non-verbal cues effectively without task-specific tuning.

Notably, even the best transcript-based LLM with Psychiatric Reasoning (GPT-4o, 60.53\%) still falls short of the 68.24\% achieved by a supervised model trained directly on transcript Qwen features, further emphasizing the performance gap between prompting-based and discriminative approaches in high-stakes diagnosis.

\begin{figure}[t]
	\centering
	\includegraphics[width=0.46\textwidth]{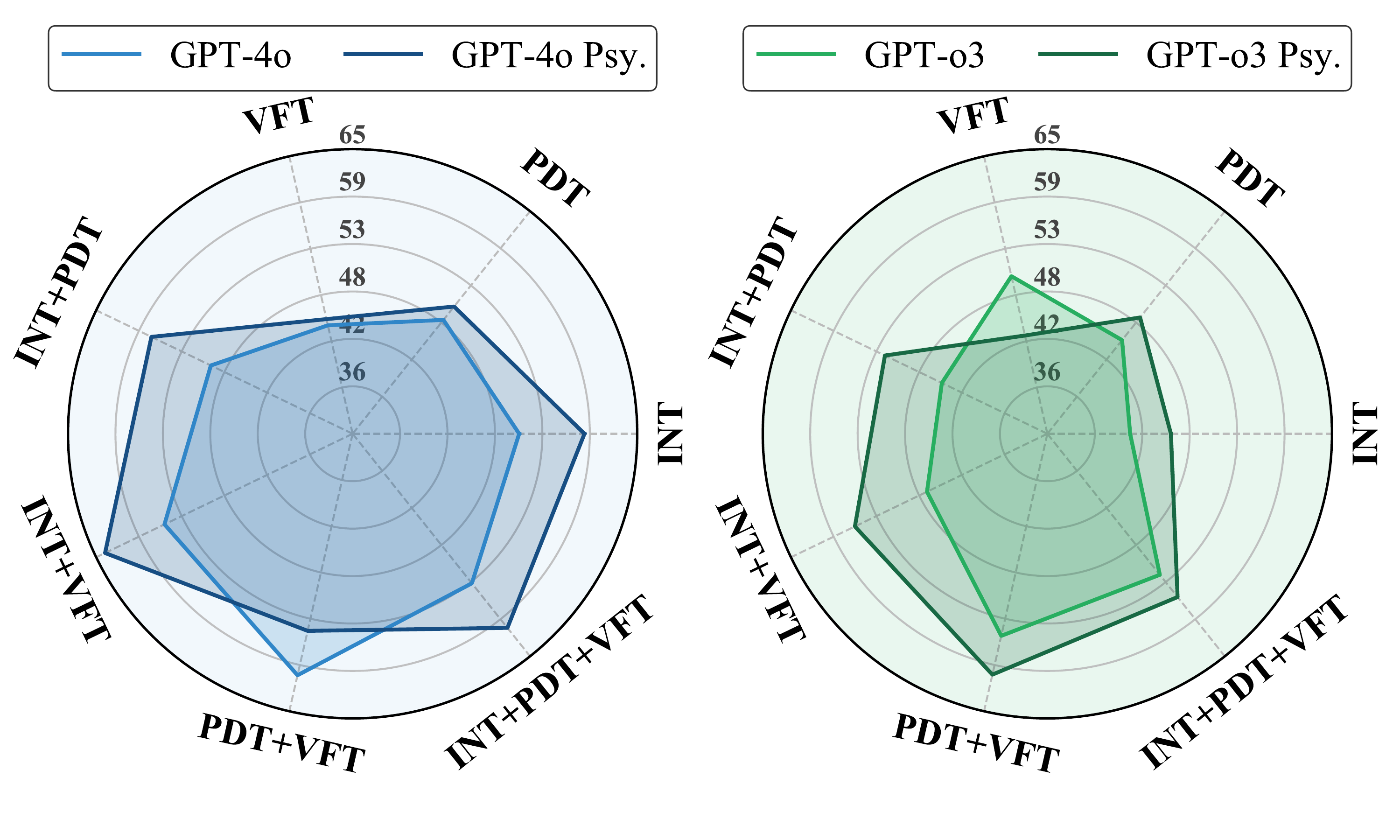}
	\caption{Performance of task combinations on LLMs.}
	\label{fig:radar}
\end{figure}

\paragraph{Task Fusion On LLMs}
To examine whether LLMs benefit from task-level information integration, we aggregate transcripts from multiple psychiatric tasks (Figure~\ref{fig:radar}). Results align with RQ1: combining tasks improves performance. For example, GPT-4o's Macro-F1 improves from 50.48\% (INT only) to 55.68\% (INT+VFT), and further to 60.53\% when Psychiatric Reasoning is applied. The best-performing configuration under transcript-only input is GPT-4o with INT+VFT and Psychiatric prompt. Similar trends are observed with GPT-o3. However, tasks like VFT remain underutilized in isolation, likely due to the loss of timing and repetition patterns during transcription. This supports our earlier findings that linguistic transcripts alone cannot fully capture the cognitive and affective richness of certain psychiatric tasks.

\subsection{Case Study}

\begin{figure}[t]
	\centering
	\hspace{-4mm}
	\includegraphics[width=0.48\textwidth]{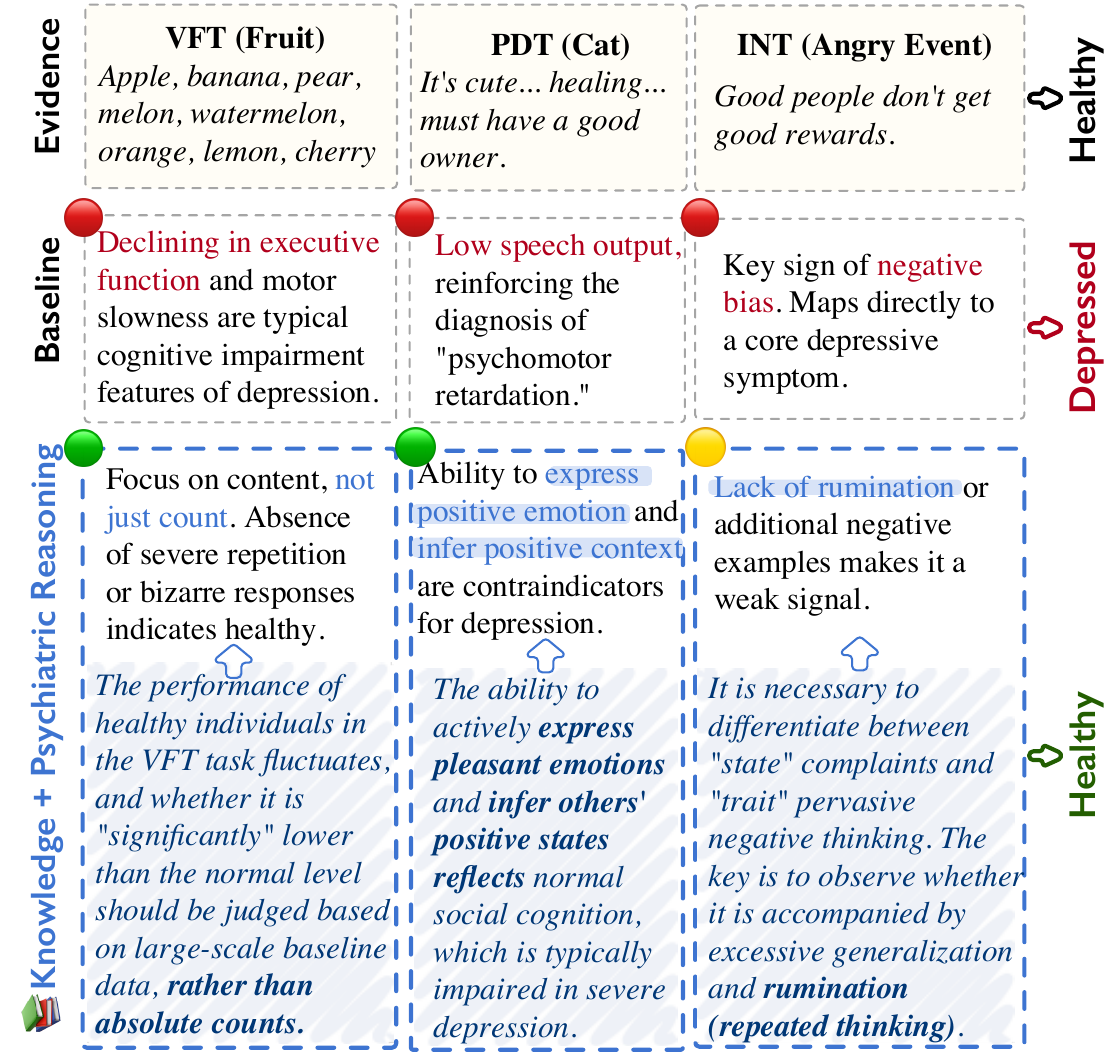}
	\caption{A case contrasting a superficial, baseline interpretation with a nuanced, knowledge-guided assessment for the same healthy participant. Red marks the incorrect assumptions of the baseline, green highlights the correct interpretations guided by psychiatric expertise, and yellow identifies points that are considered but ultimately de-emphasized.}
	\label{fig-case}
	\vspace{0em}
\end{figure}

The case study presented in Figure~\ref{fig-case} serves as a clear illustration that explicitly integrating domain-specific psychiatric knowledge enhance the performance of depression assessment.
In our observation, domain knowledge contributes in two crucial ways. 
First, it guides clinical interpretation of signals.
In the case of PDT, rather than overreacting to raw metrics \textit{``low word count''}, reasoning with psychiatric knowledge assesses recognize protective factors like emotional expressiveness, therefore correctly identifying non-pathological cases. 
Second, knowledge prevents over-weighting isolated negative signals. 
For example, when encountering \textit{``Good people don’t get good rewards,''} the baseline model treats it as a core depressive marker. 
Psychiatric reasoning, however, draws on clinical reasoning to distinguish between fleeting complaints and the pervasive negativity typical of depression. 
By noting the lack of elaboration or supporting cues, it correctly down-weights the phrase. 

\section{Related Work}
\paragraph{Corpus} The foundation of depression detection is its data corpora, which have evolved along a hierarchy of evidence, trading scale for clinical validity. 
Early research leveraged large-scale social media corpora with labels based on user self-disclosure~\cite{shen2017depression, tadesse2019detection, zirikly2019clpsych, bucur2025datasets}. 
To improve signal quality, subsequent work introduced datasets collected in controlled settings, where ground truth was typically derived from self-report questionnaire scores~\cite{gratch2014distress, valstar2016avec, guo2021deep, tasnim2022depac}. 
Representing a move toward the clinical gold standard, the most recent corpora have begun to incorporate formal diagnoses from trained psychiatrists~\cite{cai2022multi, zou2022semi, lin2022deep}.
These pioneering efforts often feature smaller or imbalanced cohorts and are focused on a limited set of behavioral tasks or modalities.
Our work therefore introduces a new clinically-validated resource featuring a balanced cohort across diverse tasks and modalities.

\paragraph{Method} Paralleling the evolution of datasets, detection methods have shifted from analyzing handcrafted features within single modalities to learning complex data representations through sophisticated, multimodal architectures.
Initial approaches relied on handcrafted features from single modalities, such as text, audio, or video~\cite{fossati2003qualitative, cummins2015speech, ma2016depaudionet}. 
A consensus has since formed around multimodal fusion models, which integrate these channels using sophisticated attention or transformer architectures to achieve stronger performance~\cite{fan2019multi, wei2023canamrf, chen-etal-2024-depression, jia2025multimodal, wu2025depmgnn}. 
The latest frontier involves applying LLMs to this task. While powerful, research highlights challenges in adapting these general-purpose models for clinical use, noting the need to imbue them with specialized, domain-specific knowledge beyond what is learned from web-scale text~\cite{guo2024large, wang-etal-2024-explainable, hua2025large, bi-etal-2025-magi}.
Our work contributes to this frontier by conducting a comprehensive analysis across different tasks and modalities to clarify their discriminative power, and then proposing a novel psychiatric reasoning mechanism to enhance the clinical awareness of LLMs.

\section{Conclusion}
We present the clinical multimodal neuropsychiatric diagnosis (C-MIND) dataset, a clinically validated resource collected from real hospital settings, featuring diverse behavioral signals across structured tasks and synchronized modalities. Through systematic analysis, we reveal how specific combinations of tasks and modalities enhance diagnostic stability, providing empirical guidance for system design. We also show that large language models, when guided by structured psychiatric knowledge, can better approximate expert reasoning in complex diagnostic scenarios. By integrating high-quality clinical data with interpretable and knowledge-informed modeling, this work offers a concrete step toward computational systems that are accurate, trustworthy, and deployable in real-world mental healthcare.

\bibliography{cmind}

\appendix

\section{C-MIND Collection Details}

\subsection{Experimental Materials}

\paragraph{1) Interview Task (INT)}
The task has a practice part and a formal part.

\noindent \textbf{Practice Session}
\begin{itemize}[leftmargin=1em, topsep=0pt, itemsep=0pt, parsep=0pt]
	\item \textit{Instructions:} ``A few questions will show up on the screen. Please say your answers. If you are ready, press the spacebar to start the practice.''
	\item \textit{Procedure:} First, a dot appears for 500ms. Then, a question appears with a note: ``You have 5 seconds to prepare.'' This screen shows for 10 seconds. Next, a new screen shows the same question with the note: ``Now, please begin your answer.''
	\item \textit{Material:} The practice question is: \texttt{Please describe one thing you regret the most.}
\end{itemize}

\noindent \textbf{Formal Session}
\begin{itemize}[leftmargin=1em, topsep=0pt, itemsep=0pt, parsep=0pt]
	\item \textit{Instructions:} ``A few questions will show up on the screen. Please say your answers. If you are ready, please press the spacebar to begin.''
	\item \textit{Procedure:} Same as the practice part.
	\item \textit{Materials:} The formal session includes three questions: \texttt{a) Please describe an event that made you the angriest.} \texttt{b) Please describe an event that you felt was the most meaningful.} \texttt{c) Please describe your favorite food.}

\end{itemize}

\begin{figure*}[t]
	\centering
	\begin{subfigure}[b]{0.22\linewidth}
		\centering
		\includegraphics[height=3cm]{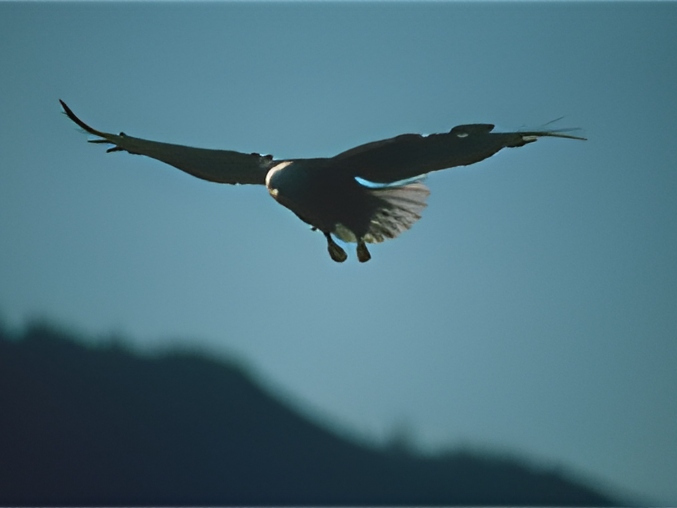}
		\caption{Image: Flying bird}
		\label{fig:bird}
	\end{subfigure}
	\hfill
	\begin{subfigure}[b]{0.22\linewidth}
		\centering
		\includegraphics[height=3cm]{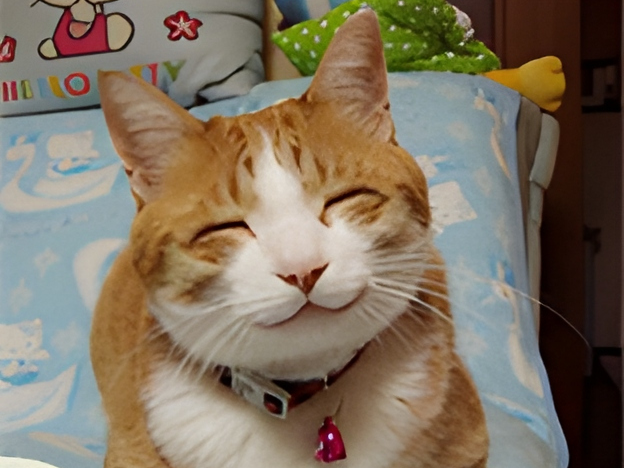}
		\caption{Image: Cat}
		\label{fig:cat}
	\end{subfigure}
	\hfill
	\begin{subfigure}[b]{0.22\linewidth}
		\centering
		\includegraphics[height=3cm]{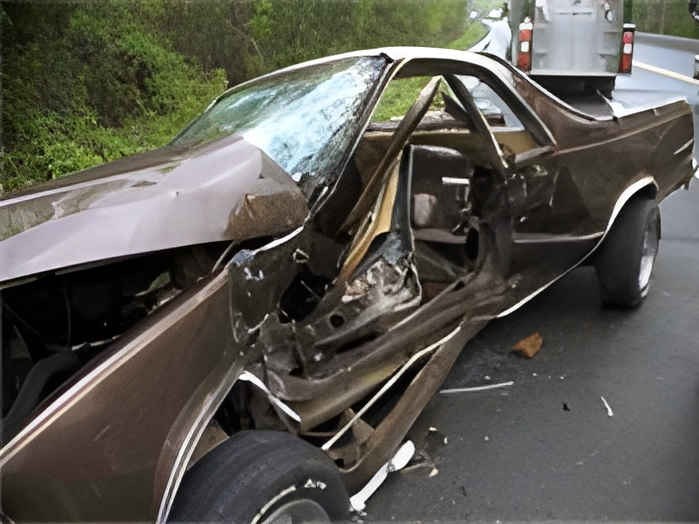}
		\caption{Image: Car accident}
		\label{fig:car_accident}
	\end{subfigure}
	\hfill
	\begin{subfigure}[b]{0.22\linewidth}
		\centering
		\includegraphics[height=3cm]{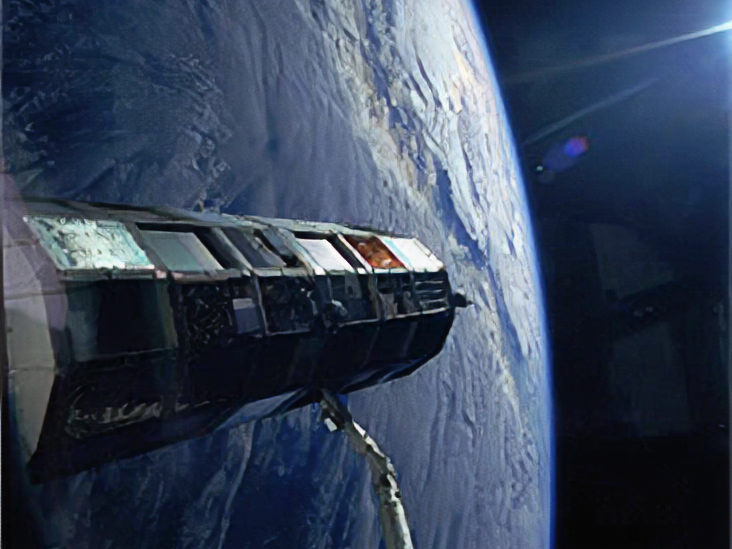}
		\caption{Image: Spaceship}
		\label{fig:spaceship}
	\end{subfigure}
	
	\caption{Visual stimuli used in the Picture Description Task (PDT). Figure (a) is the image for the practice session, while (b), (c), and (d) are the images for the formal experiment.}
	\label{fig:pdt_images}
	\vspace{-1em}
\end{figure*}

\paragraph{2) Picture Description Task (PDT)}
This task also has a practice and a formal part. The images used are shown in Figure~\ref{fig:pdt_images}.

\begin{figure*}[t!]
	\centering
	\begin{subfigure}[b]{0.48\textwidth}
		\centering
		\includegraphics[width=0.9\linewidth]{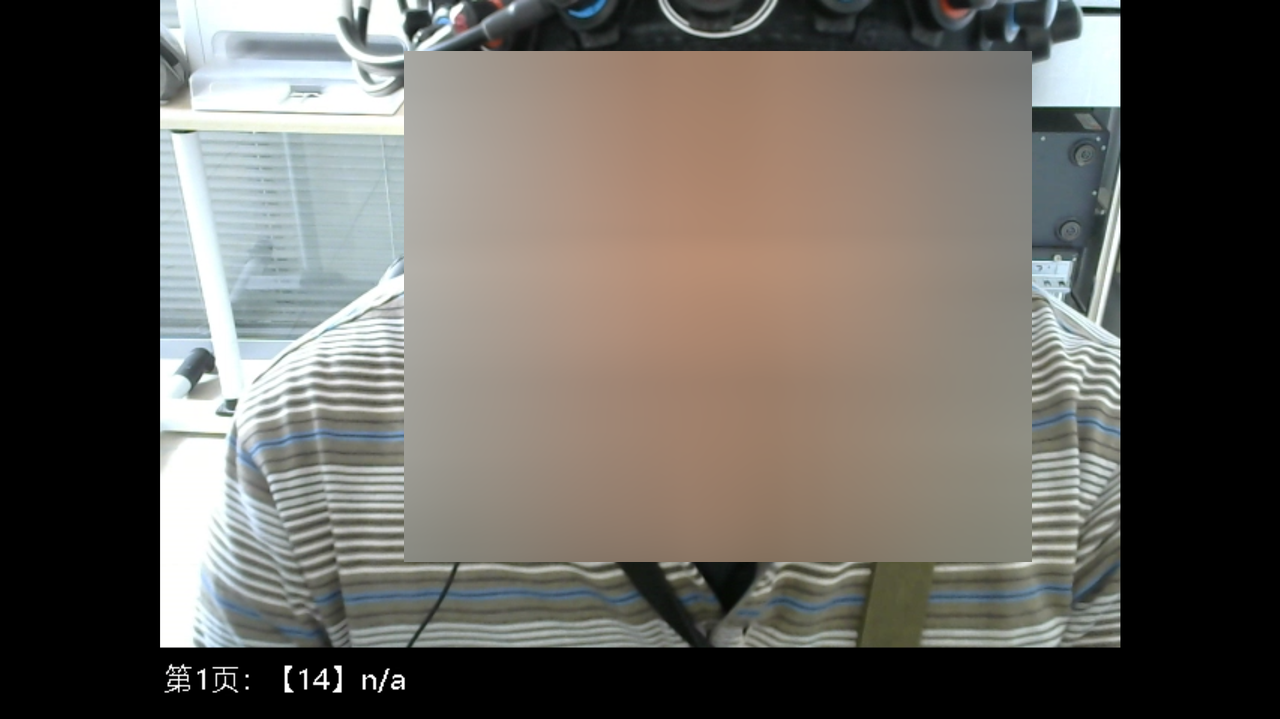}
		\caption{A male participant in the data collection environment.}
		\label{fig:male_session}
	\end{subfigure}
	\hfill 
	\begin{subfigure}[b]{0.40\textwidth}
		\centering
		\includegraphics[width=0.9\linewidth]{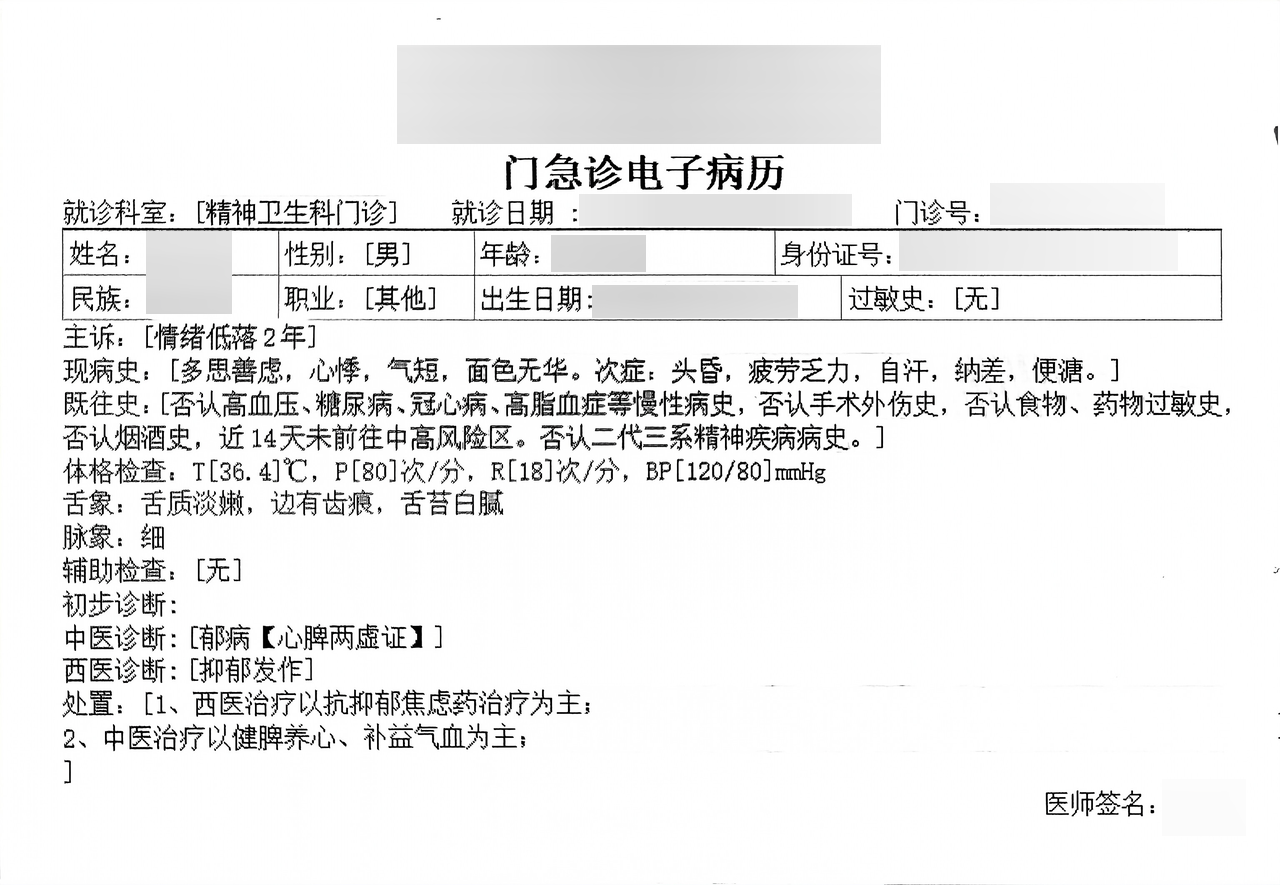}
		\caption{Anonymized clinical diagnosis record.}
		\label{fig:male_record}
	\end{subfigure}
	
	\caption{Example of a male participant.}
	\label{fig:male_example}
	\vspace{-1em}
\end{figure*}

\begin{figure*}[t!]
	\centering
	\begin{subfigure}[b]{0.48\textwidth}
		\centering
		\includegraphics[width=0.9\linewidth]{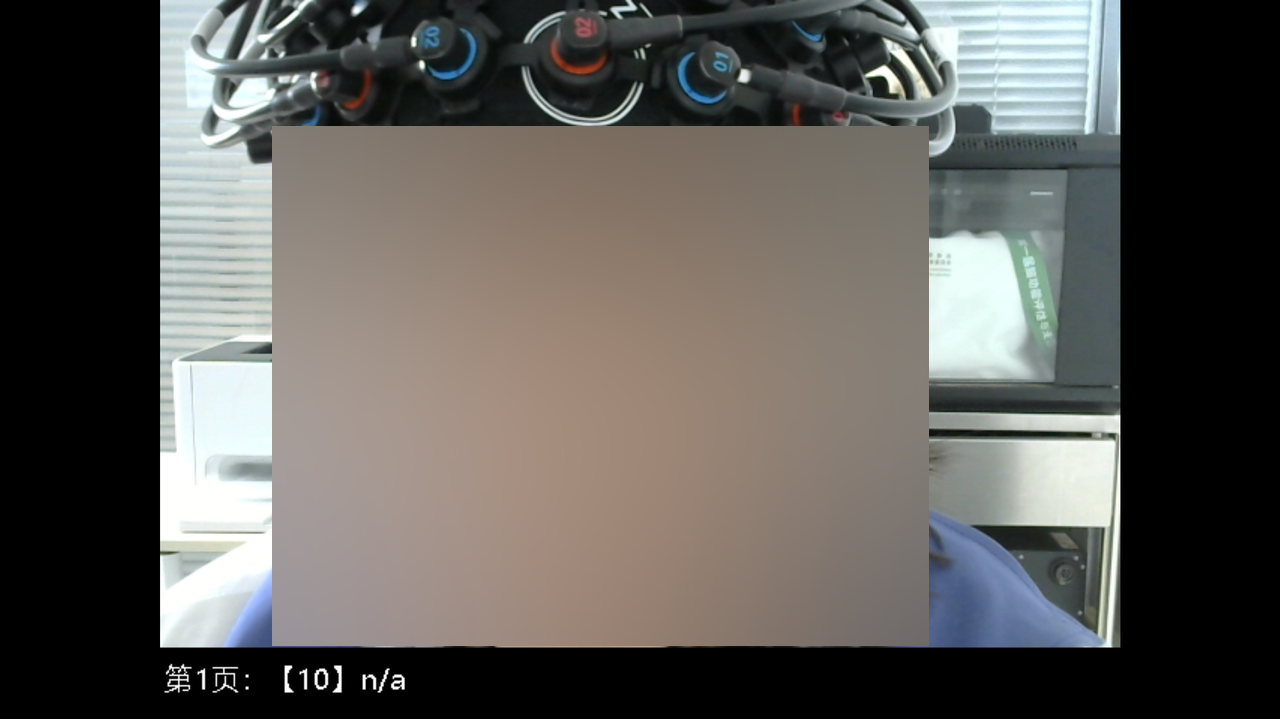}
		\caption{A female participant in the data collection environment.}
		\label{fig:female_session}
	\end{subfigure}
	\hfill 
	\begin{subfigure}[b]{0.40\textwidth}
		\centering
		\includegraphics[width=0.9\linewidth]{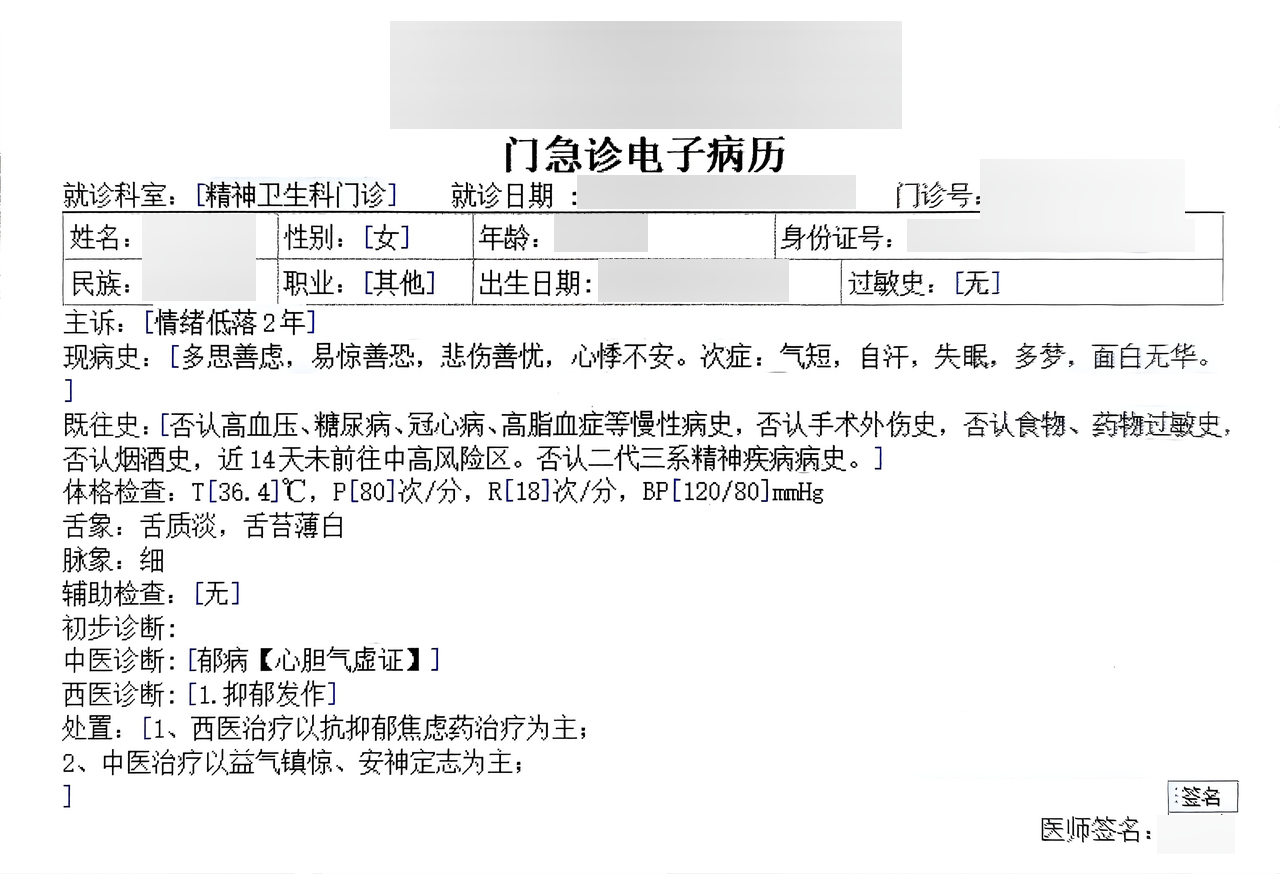}
		\caption{Anonymized clinical diagnosis record.}
		\label{fig:female_record}
	\end{subfigure}
	
	\caption{Example of a female participant. }
	\label{fig:female_example}
	\vspace{-1em}
\end{figure*}

\noindent \textbf{Practice Session}
\begin{itemize}[leftmargin=1em, topsep=0pt, itemsep=0pt, parsep=0pt]
	\item \textit{Instructions:} ``Next, you will see a picture. Please talk about the feelings or thoughts the picture gives you. Note: Please only talk about the picture itself. If you are ready, press the spacebar to start the practice.''
	\item \textit{Procedure:} After a 500ms dot, a picture appears. The text above it says: ``What thoughts or feelings does this picture give you? You have 5 seconds to prepare.'' This screen shows for 10 seconds. Then, the screen shows the same picture with new text: ``Now, please begin talking.''
	\item \textit{Material:} The practice picture is an image of a \texttt{flying bird} (see Figure~\ref{fig:bird}).
\end{itemize}
\noindent \textbf{Formal Session}
\begin{itemize}[leftmargin=1em, topsep=0pt, itemsep=0pt, parsep=0pt]
	\item \textit{Instructions:} ``Next, you will see a picture. Please talk about the feelings or thoughts the picture gives you. Note: Please only talk about the picture itself. If you are ready, press the spacebar to start the formal experiment.''
	\item \textit{Procedure:} Same as the practice part.
	\item \textit{Materials:} The formal session uses three pictures: a \texttt{cat} (Figure~\ref{fig:cat}), a \texttt{car accident} (Figure~\ref{fig:car_accident}), and a \texttt{spaceship} (Figure~\ref{fig:spaceship}).
\end{itemize}

\paragraph{3) Verbal Fluency Task (VFT)}
This task asks people to say words from a specific group.

	\noindent \textbf{Practice Session}
	\begin{itemize}[leftmargin=1em, topsep=0pt, itemsep=0pt, parsep=0pt]
		\item \textit{Instructions:} ``A word for a category will appear on the screen. Please say words from that category. You have 30 seconds. If you are ready, press the spacebar to start.''
		\item \textit{Procedure:} After a 500ms dot, a screen shows a category word with examples. The text above says: ``Now, please state your answer.'' This screen is on for 30 seconds. Then, there is a 10-second rest.
		\item \textit{Material:} The practice category is: \texttt{Green plants} (e.g., you can say ivy, cabbage, etc.).
	\end{itemize}
	\noindent \textbf{Formal Session}
	\begin{itemize}[leftmargin=1em, topsep=0pt, itemsep=0pt, parsep=0pt]
		\item \textit{Instructions:} ``A word for a category will appear on the screen. Please say words from that category. You have 30 seconds! If you are ready, please press the spacebar to start the formal experiment.''
		\item \textit{Procedure:} Same as the practice part.
		\item \textit{Materials:} The categories for the formal experiment are: \texttt{a) Four-legged animals} (e.g., pig, cow, etc.) \texttt{b) Round fruits} (e.g., watermelon, apple, etc.) \texttt{c) Cities with two-character names} (e.g., Beijing, Shanghai, etc.) \texttt{d) Vegetables} (e.g., carrot, broccoli, etc.)
	\end{itemize}

\subsection{Participant Examples}
To give a clear picture of our data, this section shows two examples: one male participant (Figure~\ref{fig:male_example}) and one female participant (Figure~\ref{fig:female_example}). For each person, we show two images. The first image shows the person in our lab during the experiment. This shows our data collection setup. The second image is a sample of the real, anonymized clinical record. This record is the gold-standard ground truth we use for the final diagnosis label in our study.

\subsection{Details of Psychometric Questionnaires}

To get a full picture of each participant’s emotional state, we use a set of standard questionnaires to help check for related factors like sleep problems or personality traits. 
The details of each instrument are as follows.

\noindent \paragraph{1) HAMD}  Hamilton Depression Rating Scale~\cite{Hamilton1960}. This is a clinician-rated tool to get an objective evaluation of depression severity. It is based on observed behaviors and what the participant reports.

\noindent \paragraph{2) HAMA}  Hamilton Anxiety Rating Scale~\cite{Hamilton1959}. This is also a clinician-rated tool that gives an objective evaluation of how severe a person’s anxiety is.

\noindent \paragraph{3) SDS} Self-Rating Depression Scale~\cite{Zung1965}. This is a self-report tool. It helps capture a person’s own feelings of depression, which might not be clear from an interview alone.

\noindent \paragraph{4) SAS} Self-Rating Anxiety Scale~\cite{Zung1971}. Like the SDS, this is a self-report tool for a person’s own feelings of anxiety.

\noindent \paragraph{5) PSQI} Pittsburgh Sleep Quality Index~\cite{Buysse1989}. This scale measures a person’s sleep quality over the last month. We include it because sleep problems are common in mood disorders.

\noindent \paragraph{6) 16PF} 16 Personality Factor Questionnaire~\cite{Cattell1970}. This tool assesses personality traits. This helps us understand if certain personality types are related to a person’s mental state.

\noindent \paragraph{7) SCL-90} Symptom Checklist-90~\cite{Derogatis1977}. This scale screens for a wide range of psychological symptoms. This helps us check for other conditions that might exist alongside depression.

\noindent \paragraph{8) HCL-32} Hypomania Checklist-32~\cite{Angst2005}. This checklist helps identify signs of hypomania. We use it to make sure we do not misdiagnose someone with bipolar disorder as having depression. This improves our diagnostic accuracy.

\section{Feature Extraction Details}

We extract two different sets of features for experiments. The first is a ``Classical Feature Set'' from common toolkits. The second is a ``Foundation Model Feature Set'' from large pretrained models.

\subsection{Classical Feature Set}

\paragraph{1) Audio} We use the eGeMAPS feature set from OpenSMILE 3.0 \cite{opensmile2010} to extract features from the audio clips. This feature set includes statistics for 8 frequency-related features, 3 energy-related features, and 14 spectral features. This gives a total of 88 features for audio.

\paragraph{2) Video} We use OpenFace 2.2.0 \cite{Baltrusaitis2018} to extract facial features from video files. OpenFace is an open-source tool for facial analysis. We extract features like frame number, confidence, eye gaze vectors, eye landmarks, head pose, and facial action units (AUs). We then take frames with a confidence score greater than 0.75 and use MATLAB’s \texttt{smooth} function to smooth the data with a window size of 11. For all processed features, we calculate 7 statistics: minimum, maximum, mean, variance, range, kurtosis, and skewness. This gives a total of 4,963 features.

\paragraph{3) Transcript} We use a Chinese pretrained DeBERTa model \cite{he2020deberta} to extract 768-dimension feature vector for the text.

\paragraph{4) fNIRS} We collect the oxygenated and deoxygenated hemoglobin concentration from 45 channels of the fNIRS device \cite{cui2010fnirs}. For this data, we calculate the same 7 statistics as the video features: minimum, maximum, mean, variance, range, kurtosis, and skewness. This gives a total of 630 features for fNIRS.

\subsection{Foundation Model Feature Set}
We use large Qwen models to create another feature set. All models run using vLLM for deployment.

\paragraph{1) Audio} We use \texttt{Qwen2-Audio-7B-Instruct} \cite{chu2023qwen}. We take the audio for each of the 10 tasks and cut it into clips shorter than 30 seconds. We feed these clips into the model and take the features from the last hidden layer. We join the features for the same task and then apply global max pooling over the time dimension. This gives a final 4,096-dimension audio feature.

\paragraph{2) Video} We use \texttt{Qwen2.5-VL-72B} \cite{bai2025qwen2}. First, we split each participant’s full video into 10 smaller videos based on the 10 task timestamps. We feed each video into the model. Each frame is resized to 224x224 pixels. We use the official \texttt{get\_video\_features} function to get the feature tensor. Then, we apply global max pooling over the time dimension. This gives a final 8,192-dimension video feature.

\paragraph{3) Transcript} We use \texttt{Qwen3-235B-A22B} \cite{yang2025qwen3}. We take the speech-to-text transcripts for each of the 10 tasks and feed them into the model. We extract features from the last hidden layer and then apply global max pooling over the sequence dimension. This gives a final 4,096-dimension text feature.

\paragraph{4) fNIRS} Due to the lack of a pretrained model for fNIRS signals, we use the same statistical features for fNIRS as in the Classical Feature Set.

\section{Methodology Implementation Details}

\subsection{Behavioral Signature Modeling}
We train six different learning backbones to model the behavioral signatures. The implementation details for each model are as follows:

\paragraph{1) LSTM} For the LSTM model, we first extract features for specific time steps based on the task type and normalize the data for each modality. Each modality passes through a Batch Normalization (BN) layer, followed by a two-layer fully-connected network for dimension reduction (hidden dimensions 512 then 128), with a Dropout rate of 0.2. For single-modality cases, the reduced features are used directly. For multi-modality, all modal features are first concatenated. Then, global average pooling is used to generate one-dimensional gate weights, which are used to fuse the information from different paths in a weighted manner. The fused features then enter a two-layer bidirectional LSTM, with each layer containing 512 hidden units and a Dropout rate of 0.3. This is followed by a single-head self-attention layer with a key dimension of 128 and a dropout rate of 0.3. Afterwards, we apply global average pooling, feed the result into a 128-unit fully-connected layer, and finally use a 1-unit Sigmoid layer to output the binary classification probability. We use the AdamW optimizer for training with a learning rate of 1e-4, weight decay of 1e-4, and a gradient clipping norm of 1.0 with amsgrad enabled. The loss function is binary cross-entropy. We monitor the validation loss and use an early stopping mechanism with a patience of 20 epochs, saving the best model from each training run.

\paragraph{2) CNN} To fit the CNN’s input structure, the 512-dimensional fused features are reshaped into a 3D tensor of shape (512, 1, 1). The model first receives this input tensor and passes it through a convolutional layer with 32 filters of size (2, 1), a stride of (2, 1), and ‘same’ padding. This is followed by a batch normalization layer and a ReLU activation function. To mitigate overfitting, a Dropout layer with a rate of 0.3 is added. Next, a Flatten layer is used to flatten the multi-dimensional output, which is then fed into a fully-connected layer with 32 neurons, a ReLU activation, and L2 regularization ($\lambda=0.01$). Another Dropout layer (rate 0.3) follows. The final output layer has a single neuron with a sigmoid activation function to produce the probability for the positive class. The model is trained using the SGD optimizer with a learning rate of 1e-4 and binary cross-entropy as the loss function. We use callbacks for EarlyStopping, ReduceLROnPlateau, and ModelCheckpoint. The model with the best validation performance is saved.

\paragraph{3) MLP} The Multi-Layer Perceptron model has two hidden layers. The first is a fully-connected layer with 64 neurons, followed by a ReLU activation and Dropout (rate 0.3). The second is a 32-neuron fully-connected layer with the same structure. To enhance generalization, we also add an L2 regularization term ($\lambda = 0.01$) to both hidden layers. The output layer uses a Sigmoid activation function for the binary classification task. The model is trained using the SGD optimizer with an initial learning rate of 0.001. It also uses callbacks like EarlyStopping, ReduceLROnPlateau, and ModelCheckpoint to dynamically adjust the learning rate and prevent overfitting. During training, the validation loss is monitored, and the model parameters with the best validation performance are saved.

\paragraph{4) k-NN} For the K-Nearest Neighbors model, we perform an automatic selection of the hyperparameter $k$. We test different numbers of neighbors from 1 to 10 and use five-fold cross-validation to evaluate the accuracy for each. The $k$ value that yields the highest average accuracy is selected to build the final model.

\paragraph{5) RF} For the Random Forest model, we use RandomizedSearchCV with five-fold cross-validation to search for the best hyperparameters. We search over two key parameters: the number of trees ($n\_estimators$, from 25 to 500) and the maximum depth of the trees ($max\_depth$, from 2 to 50). The final model is then rebuilt, trained, and evaluated using the best parameters found in the search.

\begin{table*}[t]
	\centering
	
	\renewcommand{\arraystretch}{1.0}
	\setlength{\tabcolsep}{0.7mm}
	\resizebox{1.\textwidth}{!}{
		\begin{tabular}{lccccccccccc}
			\toprule
			\textbf{Model} & \textbf{Version} & \textbf{Temperature} & \textbf{Top P} & \textbf{Frequency Penalty} & \textbf{Presence Penalty} & \textbf{Stop} & \textbf{Logprobs} & \textbf{Stream} & \textbf{Logit Bias} & \textbf{n} \\
			\midrule
			GPT-4o & 2024-05-13 & 0 & 1.0 & 0 & 0 & none & false & false & none & 1 \\
			GPT-o3 & 2025-04-16 & 0 & 1.0 & 0 & 0 & none & false & false & none & 1 \\
			DeepSeek-R1 & 2025-01-20 & 0 & 0.95 & 0 & 0 & none & false & false & none & 1 \\
			DeepSeek-V3 & 2024-12-26 & 0 & 0.95 & 0 & 0 & none & false & false & none & 1 \\
			Qwen3 (NT) & - & 0 & 1.0 & 0 & 0 & none & false & - & none & 1 \\
			Qwen3 (T) & - & 0 & 1.0 & 0 & 0 & none & false & - & none & 1 \\
			Qwen2.5-Omni & - & 0 & 1.0 & 0 & 0 & none & false & - & none & 1 \\
			\bottomrule
		\end{tabular}
	}
	\caption{API parameters for LLM inference.}
	\vspace{0em}
	\label{tab:llm_params}
\end{table*}

\begin{table*}[h!] 
	\centering 
	\begin{small} 
		
		\renewcommand{\arraystretch}{1.}
		\begin{tabular}{|p{1.0\textwidth}|}
			\hline
			\textbf{[Task Setting]} \newline
			A depression detection task is required. Based on the speech-to-text content of subjects in experimental tasks, determine whether they are depression patients (1) or healthy controls (0). This experiment aims to study the linguistic features of subjects under different tasks to distinguish between depression patients and healthy controls. Each subject completes 10 tasks of the following types:
			\textbf{1) Interview Tasks (T1-T3):} Subjects answer questions about ``an anger-inducing event,'' ``a meaningful event,'' and ``their favorite food.''
			\textbf{2) Picture Description Tasks (T4-T6):} Subjects describe pictures of a ``cat,'' a ``car accident,'' and a ``spaceship,'' and engage in free association.
			\textbf{3) Verbal Fluency Tasks (T7-T10):} Subjects perform free association based on the categories: ``four-legged things,'' ``fruits,'' ``cities,'' and ``vegetables.''
			If a subject’s content for a specific task was not collected, it is marked as ``\$missing''. \\
			\hline
			\textcolor{blue}{\textsc{Only in Few-Shot Settings}} \newline
			\textbf{[Example Samples]} \newline
			\textit{\{\texttt{Depressed Sample}\}}  $+$
			\textit{\{\texttt{Control Sample}\}} \\
			\hline
			\textbf{[Sample to be Judged]} \newline
			\textit{\{\texttt{Test Sample}\}} \\
			\hline
			\textcolor{blue}{\textsc{Only for Psychiatric Reasoning}} \newline
			\textbf{[Reasoning Process]} \newline
			Think based on the following diagnostic clues:
			The three types of tasks have their own testing goals. The interview tasks mainly test the subject’s emotional response and self-awareness. The picture description tasks mainly test semantic fluency, imagination, and cognitive flexibility. The verbal fluency tasks are mainly used to evaluate verbal fluency and cognitive flexibility. \newline
			\textbf{For patients with depression,} they tend to use more negative words and lack detailed descriptions in interview tasks. In picture description tasks, their descriptions are simpler, lean towards negative outcomes, and lack emotional expression. In verbal fluency tasks, they list fewer items and may have repetitions or words irrelevant to the topic. \newline
			\textbf{For healthy subjects,} they tend to be more rational and specific in their descriptions in interview tasks. In picture description tasks, their descriptions are more detailed, their reflections on events are more diverse, and they show richer emotional expression. In verbal fluency tasks, they tend to list a richer variety of items. \\
			\hline
			\textbf{[Task Requirements]} \newline
			\textcolor{blue}{\textsc{For Zero-Shot Settings}} \newline
			Your task is to determine whether the subject is a depression patient (labeled 1) or a healthy control (labeled 0) based on the text content from [T1] to [T10]. A judgment must be made regardless of how many ``\$missing'' entries are present. \newline
			\textcolor{blue}{\textsc{For Few-Shot Settings}} \newline
			Your task is to determine whether the subject is a depression patient (labeled 1) or a healthy control (labeled 0) based on the text content from [T1] to [T10] in the ``Sample to be Judged,'' in conjunction with the ``Example Samples.'' A judgment must be made regardless of how many ``\$missing'' entries are present. \\
			\hline
			\textbf{[Output Requirements]} \newline
			\textcolor{blue}{\textsc{For Direct Method}} \newline
			If judged as depression, output ``{{1}}''. If judged as a healthy control, output ``{{0}}''. Do not output any text other than ``{{1}}'' or ``{{0}}''. Do not output any other explanatory content. \newline
			\textcolor{blue}{\textsc{For Zero-Shot Vanilla/Psychiatric Reasoning}} \newline
			First, output the reasoning process, and then on a new line, output only the judgment result. If judged as depression, output ``{{1}}''; if judged as a healthy control, output ``{{0}}''. \newline
			\textcolor{blue}{\textsc{For Few-Shot Vanilla/Psychiatric Reasoning}} \newline
			First, output the reasoning process, summarizing and comparing the depression patient and healthy control in the ``Example Samples.'' Then, on a new line, output only the judgment result. If judged as depression, output ``{{1}}''; if judged as a healthy control, output ``{{0}}''. \\
			\hline
		\end{tabular}
	\end{small}
	\caption{Unified prompt template for depression assessment.} 
	\label{tab:prompt_template} 
\end{table*}

\paragraph{6) SVM} For the Support Vector Machine model, we use a linear kernel. The regularization strength $C$ is set to 1, and we set $class\_weight=balanced$ to address the class imbalance issue in the dataset.

\subsection{Psychiatric Reasoning with LLMs}
We evaluate several leading LLMs to assess their psychiatric reasoning capabilities. We select seven models for our experiments, including text-based and multimodal models. The models and their specific API parameters used for inference are described below.

\paragraph{1) GPT-4o}: A model from OpenAI, released on May 13, 2024. It can process any combination of text, image, and video as input and generate text, audio, and image outputs~\cite{openai2024gpt4o}.

\paragraph{2) GPT-o3}: A model from OpenAI, released on April 16, 2025. It is described as OpenAI’s most powerful reasoning model, advancing capabilities in coding, math, and visual perception~\cite{openai2025o3}.

\paragraph{3) DeepSeek-R1}: A reasoning model from the AI company DeepSeek, released on January 20, 2025. Its model weights are open-sourced~\cite{guo2025deepseek}.

\paragraph{4) DeepSeek-V3}: An LLM from DeepSeek AI, released and open-sourced on December 26, 2024~\cite{liu2024deepseek}.

\paragraph{5) Qwen3-235B-A22B}: A model from Alibaba, released on April 29, 2025. A key innovation is integrating a ``thinking mode'' for complex reasoning with a ``non-thinking mode'' for fast responses into one framework. We test both modes, referred to as Qwen3 (T) and Qwen3 (NT)~\cite{yang2025qwen3}.

\paragraph{6) Qwen2.5-Omni}: An end-to-end omni-modal model from the Qwen team, released on March 27, 2025. It supports simultaneous input of text, image, audio, and video with real-time streaming output~\cite{xu2025qwen2}.

All models are called using their official APIs with fixed parameters to ensure deterministic outputs. The prompts for direct prediction, vanilla reasoning and psychiatric reasoning are shown in Table \ref{tab:prompt_template}. The specific parameters for each model are detailed in Table~\ref{tab:llm_params}.

\begin{table*}[htbp]
	\setlength{\tabcolsep}{0.7mm}
	\small
	\centering
	\renewcommand{\arraystretch}{1.6}
	\centering
	\resizebox{\textwidth}{!}{%
		\begin{tabular}{lllcccccccccccccccccc}
			\toprule
			& & & \multicolumn{6}{c}{\textbf{Interview (INT)}} & \multicolumn{6}{c}{\textbf{Picture Description (PDT)}} & \multicolumn{6}{c}{\textbf{Verbal Fluency (VFT)}} \\
			\cmidrule(lr){4-9} \cmidrule(lr){10-15} \cmidrule(lr){16-21}
			\textbf{Modality} & \textbf{Feature Set} & \textbf{Metric} & \textbf{LSTM} & \textbf{CNN} & \textbf{MLP} & \textbf{k-NN} & \textbf{RF} & \textbf{SVM} & \textbf{LSTM} & \textbf{CNN} & \textbf{MLP} & \textbf{k-NN} & \textbf{RF} & \textbf{SVM} & \textbf{LSTM} & \textbf{CNN} & \textbf{MLP} & \textbf{k-NN} & \textbf{RF} & \textbf{SVM} \\
			\midrule
			
			\multirow{12}{*}{\textbf{Audio}} & \multirow{6}{*}{OpenSmile} 
			& Macro-F1             & \cellcolor[HTML]{B8DAE9}72.15 & \cellcolor[HTML]{B8DAE9}84.25 & \cellcolor[HTML]{B8DAE9}82.31 & \cellcolor[HTML]{B8DAE9}85.18 & \cellcolor[HTML]{B8DAE9}91.17 & \cellcolor[HTML]{B8DAE9}91.11 & \cellcolor[HTML]{6BAED6}69.49 & \cellcolor[HTML]{6BAED6}88.24 & \cellcolor[HTML]{6BAED6}81.21 & \cellcolor[HTML]{6BAED6}94.10 & \cellcolor[HTML]{6BAED6}88.24 & \cellcolor[HTML]{6BAED6}85.18 & \cellcolor[HTML]{7FB8DD}70.90 & \cellcolor[HTML]{7FB8DD}84.28 & \cellcolor[HTML]{7FB8DD}79.39 & \cellcolor[HTML]{7FB8DD}76.14 & \cellcolor[HTML]{7FB8DD}88.24 & \cellcolor[HTML]{7FB8DD}82.29 \\
			& & Accuracy                & \cellcolor[HTML]{B8DAE9}72.35 & \cellcolor[HTML]{B8DAE9}84.31 & \cellcolor[HTML]{B8DAE9}82.35 & \cellcolor[HTML]{B8DAE9}85.29 & \cellcolor[HTML]{B8DAE9}91.18 & \cellcolor[HTML]{B8DAE9}91.18 & \cellcolor[HTML]{6BAED6}70.00 & \cellcolor[HTML]{6BAED6}88.24 & \cellcolor[HTML]{6BAED6}81.37 & \cellcolor[HTML]{6BAED6}94.12 & \cellcolor[HTML]{6BAED6}88.24 & \cellcolor[HTML]{6BAED6}85.29 & \cellcolor[HTML]{7FB8DD}71.18 & \cellcolor[HTML]{7FB8DD}84.31 & \cellcolor[HTML]{7FB8DD}79.41 & \cellcolor[HTML]{7FB8DD}76.47 & \cellcolor[HTML]{7FB8DD}88.24 & \cellcolor[HTML]{7FB8DD}82.35 \\
			& & Depression-P & \cellcolor[HTML]{B8DAE9}72.08 & \cellcolor[HTML]{B8DAE9}80.76 & \cellcolor[HTML]{B8DAE9}80.08 & \cellcolor[HTML]{B8DAE9}86.43 & \cellcolor[HTML]{B8DAE9}91.32 & \cellcolor[HTML]{B8DAE9}92.50 & \cellcolor[HTML]{6BAED6}66.18 & \cellcolor[HTML]{6BAED6}88.24 & \cellcolor[HTML]{6BAED6}78.22 & \cellcolor[HTML]{6BAED6}94.74 & \cellcolor[HTML]{6BAED6}88.24 & \cellcolor[HTML]{6BAED6}86.43 & \cellcolor[HTML]{7FB8DD}70.19 & \cellcolor[HTML]{7FB8DD}83.26 & \cellcolor[HTML]{7FB8DD}79.86 & \cellcolor[HTML]{7FB8DD}78.02 & \cellcolor[HTML]{7FB8DD}88.24 & \cellcolor[HTML]{7FB8DD}82.81 \\
			& & Depression-R & \cellcolor[HTML]{B8DAE9}74.12 & \cellcolor[HTML]{B8DAE9}90.20 & \cellcolor[HTML]{B8DAE9}86.27 & \cellcolor[HTML]{B8DAE9}85.29 & \cellcolor[HTML]{B8DAE9}91.18 & \cellcolor[HTML]{B8DAE9}91.18 & \cellcolor[HTML]{6BAED6}82.35 & \cellcolor[HTML]{6BAED6}88.24 & \cellcolor[HTML]{6BAED6}88.24 & \cellcolor[HTML]{6BAED6}94.12 & \cellcolor[HTML]{6BAED6}88.24 & \cellcolor[HTML]{6BAED6}85.29 & \cellcolor[HTML]{7FB8DD}76.47 & \cellcolor[HTML]{7FB8DD}86.27 & \cellcolor[HTML]{7FB8DD}78.43 & \cellcolor[HTML]{7FB8DD}76.47 & \cellcolor[HTML]{7FB8DD}88.24 & \cellcolor[HTML]{7FB8DD}82.35 \\
			& & Depression-F1      & \cellcolor[HTML]{B8DAE9}72.59 & \cellcolor[HTML]{B8DAE9}85.18 & \cellcolor[HTML]{B8DAE9}83.01 & \cellcolor[HTML]{B8DAE9}83.87 & \cellcolor[HTML]{B8DAE9}90.91 & \cellcolor[HTML]{B8DAE9}91.89 & \cellcolor[HTML]{6BAED6}73.35 & \cellcolor[HTML]{6BAED6}88.24 & \cellcolor[HTML]{6BAED6}82.75 & \cellcolor[HTML]{6BAED6}93.75 & \cellcolor[HTML]{6BAED6}88.24 & \cellcolor[HTML]{6BAED6}86.49 & \cellcolor[HTML]{7FB8DD}72.93 & \cellcolor[HTML]{7FB8DD}84.63 & \cellcolor[HTML]{7FB8DD}79.08 & \cellcolor[HTML]{7FB8DD}78.95 & \cellcolor[HTML]{7FB8DD}88.24 & \cellcolor[HTML]{7FB8DD}83.33 \\
			& & Control-F1         & \cellcolor[HTML]{B8DAE9}71.72 & \cellcolor[HTML]{B8DAE9}83.32 & \cellcolor[HTML]{B8DAE9}81.62 & \cellcolor[HTML]{B8DAE9}86.49 & \cellcolor[HTML]{B8DAE9}91.43 & \cellcolor[HTML]{B8DAE9}90.32 & \cellcolor[HTML]{6BAED6}65.63 & \cellcolor[HTML]{6BAED6}88.24 & \cellcolor[HTML]{6BAED6}79.66 & \cellcolor[HTML]{6BAED6}94.44 & \cellcolor[HTML]{6BAED6}88.24 & \cellcolor[HTML]{6BAED6}83.87 & \cellcolor[HTML]{7FB8DD}68.87 & \cellcolor[HTML]{7FB8DD}83.94 & \cellcolor[HTML]{7FB8DD}79.71 & \cellcolor[HTML]{7FB8DD}73.33 & \cellcolor[HTML]{7FB8DD}88.24 & \cellcolor[HTML]{7FB8DD}81.25 \\
			\cmidrule(lr){2-21}
			& \multirow{6}{*}{Qwen-Audio} 
			& Macro-F1             & \cellcolor[HTML]{B8DAE9}72.57 & \cellcolor[HTML]{B8DAE9}78.41 & \cellcolor[HTML]{B8DAE9}58.39 & \cellcolor[HTML]{B8DAE9}55.54 & \cellcolor[HTML]{B8DAE9}69.26 & \cellcolor[HTML]{B8DAE9}76.39 & \cellcolor[HTML]{6BAED6}74.07 & \cellcolor[HTML]{6BAED6}71.54 & \cellcolor[HTML]{6BAED6}67.58 & \cellcolor[HTML]{6BAED6}69.64 & \cellcolor[HTML]{6BAED6}72.47 & \cellcolor[HTML]{6BAED6}79.39 & \cellcolor[HTML]{7FB8DD}76.93 & \cellcolor[HTML]{7FB8DD}82.33 & \cellcolor[HTML]{7FB8DD}61.63 & \cellcolor[HTML]{7FB8DD}78.96 & \cellcolor[HTML]{7FB8DD}76.43 & \cellcolor[HTML]{7FB8DD}76.39 \\
			& & Accuracy                & \cellcolor[HTML]{B8DAE9}72.94 & \cellcolor[HTML]{B8DAE9}78.43 & \cellcolor[HTML]{B8DAE9}58.82 & \cellcolor[HTML]{B8DAE9}55.88 & \cellcolor[HTML]{B8DAE9}69.61 & \cellcolor[HTML]{B8DAE9}76.47 & \cellcolor[HTML]{6BAED6}74.12 & \cellcolor[HTML]{6BAED6}71.57 & \cellcolor[HTML]{6BAED6}68.63 & \cellcolor[HTML]{6BAED6}70.59 & \cellcolor[HTML]{6BAED6}72.55 & \cellcolor[HTML]{6BAED6}79.41 & \cellcolor[HTML]{7FB8DD}77.06 & \cellcolor[HTML]{7FB8DD}82.35 & \cellcolor[HTML]{7FB8DD}62.75 & \cellcolor[HTML]{7FB8DD}79.41 & \cellcolor[HTML]{7FB8DD}76.47 & \cellcolor[HTML]{7FB8DD}76.47 \\
			& & Depression-P & \cellcolor[HTML]{B8DAE9}71.83 & \cellcolor[HTML]{B8DAE9}77.30 & \cellcolor[HTML]{B8DAE9}57.24 & \cellcolor[HTML]{B8DAE9}56.07 & \cellcolor[HTML]{B8DAE9}69.92 & \cellcolor[HTML]{B8DAE9}76.84 & \cellcolor[HTML]{6BAED6}74.71 & \cellcolor[HTML]{6BAED6}72.22 & \cellcolor[HTML]{6BAED6}64.74 & \cellcolor[HTML]{6BAED6}73.52 & \cellcolor[HTML]{6BAED6}72.83 & \cellcolor[HTML]{6BAED6}79.51 & \cellcolor[HTML]{7FB8DD}76.70 & \cellcolor[HTML]{7FB8DD}79.92 & \cellcolor[HTML]{7FB8DD}69.47 & \cellcolor[HTML]{7FB8DD}82.20 & \cellcolor[HTML]{7FB8DD}76.66 & \cellcolor[HTML]{7FB8DD}76.84 \\
			& & Depression-R & \cellcolor[HTML]{B8DAE9}76.47 & \cellcolor[HTML]{B8DAE9}80.39 & \cellcolor[HTML]{B8DAE9}68.63 & \cellcolor[HTML]{B8DAE9}55.88 & \cellcolor[HTML]{B8DAE9}69.61 & \cellcolor[HTML]{B8DAE9}76.47 & \cellcolor[HTML]{6BAED6}72.94 & \cellcolor[HTML]{6BAED6}70.59 & \cellcolor[HTML]{6BAED6}84.31 & \cellcolor[HTML]{6BAED6}70.59 & \cellcolor[HTML]{6BAED6}72.55 & \cellcolor[HTML]{6BAED6}79.41 & \cellcolor[HTML]{7FB8DD}78.82 & \cellcolor[HTML]{7FB8DD}86.27 & \cellcolor[HTML]{7FB8DD}50.98 & \cellcolor[HTML]{7FB8DD}79.41 & \cellcolor[HTML]{7FB8DD}76.47 & \cellcolor[HTML]{7FB8DD}76.47 \\
			& & Depression-F1      & \cellcolor[HTML]{B8DAE9}73.17 & \cellcolor[HTML]{B8DAE9}78.76 & \cellcolor[HTML]{B8DAE9}62.36 & \cellcolor[HTML]{B8DAE9}51.61 & \cellcolor[HTML]{B8DAE9}66.58 & \cellcolor[HTML]{B8DAE9}77.78 & \cellcolor[HTML]{6BAED6}73.69 & \cellcolor[HTML]{6BAED6}71.34 & \cellcolor[HTML]{6BAED6}72.99 & \cellcolor[HTML]{6BAED6}64.29 & \cellcolor[HTML]{6BAED6}71.43 & \cellcolor[HTML]{6BAED6}80.00 & \cellcolor[HTML]{7FB8DD}77.44 & \cellcolor[HTML]{7FB8DD}82.96 & \cellcolor[HTML]{7FB8DD}57.37 & \cellcolor[HTML]{7FB8DD}75.86 & \cellcolor[HTML]{7FB8DD}75.91 & \cellcolor[HTML]{7FB8DD}77.78 \\
			& & Control-F1         & \cellcolor[HTML]{B8DAE9}71.96 & \cellcolor[HTML]{B8DAE9}78.06 & \cellcolor[HTML]{B8DAE9}54.41 & \cellcolor[HTML]{B8DAE9}59.46 & \cellcolor[HTML]{B8DAE9}71.93 & \cellcolor[HTML]{B8DAE9}75.00 & \cellcolor[HTML]{6BAED6}74.45 & \cellcolor[HTML]{6BAED6}71.75 & \cellcolor[HTML]{6BAED6}62.18 & \cellcolor[HTML]{6BAED6}75.00 & \cellcolor[HTML]{6BAED6}73.52 & \cellcolor[HTML]{6BAED6}78.79 & \cellcolor[HTML]{7FB8DD}76.41 & \cellcolor[HTML]{7FB8DD}81.69 & \cellcolor[HTML]{7FB8DD}65.89 & \cellcolor[HTML]{7FB8DD}82.05 & \cellcolor[HTML]{7FB8DD}76.95 & \cellcolor[HTML]{7FB8DD}75.00 \\
			\midrule
			
			\multirow{12}{*}{\textbf{Video}} & \multirow{6}{*}{OpenFace} 
			& Macro-F1             & \cellcolor[HTML]{A8D1E4}70.41 & \cellcolor[HTML]{A8D1E4}71.47 & \cellcolor[HTML]{A8D1E4}69.00 & \cellcolor[HTML]{A8D1E4}72.94 & \cellcolor[HTML]{A8D1E4}72.40 & \cellcolor[HTML]{A8D1E4}73.32 & \cellcolor[HTML]{9ECAE1}72.25 & \cellcolor[HTML]{9ECAE1}75.46 & \cellcolor[HTML]{9ECAE1}65.42 & \cellcolor[HTML]{9ECAE1}69.64 & \cellcolor[HTML]{9ECAE1}74.49 & \cellcolor[HTML]{9ECAE1}64.58 & \cellcolor[HTML]{C6DBEF}74.64 & \cellcolor[HTML]{C6DBEF}74.49 & \cellcolor[HTML]{C6DBEF}69.54 & \cellcolor[HTML]{C6DBEF}69.64 & \cellcolor[HTML]{C6DBEF}75.43 & \cellcolor[HTML]{C6DBEF}64.71 \\
			& & Accuracy                & \cellcolor[HTML]{A8D1E4}70.59 & \cellcolor[HTML]{A8D1E4}71.57 & \cellcolor[HTML]{A8D1E4}69.61 & \cellcolor[HTML]{A8D1E4}73.53 & \cellcolor[HTML]{A8D1E4}72.55 & \cellcolor[HTML]{A8D1E4}73.53 & \cellcolor[HTML]{9ECAE1}72.35 & \cellcolor[HTML]{9ECAE1}75.49 & \cellcolor[HTML]{9ECAE1}65.69 & \cellcolor[HTML]{9ECAE1}70.59 & \cellcolor[HTML]{9ECAE1}74.51 & \cellcolor[HTML]{9ECAE1}64.71 & \cellcolor[HTML]{C6DBEF}74.71 & \cellcolor[HTML]{C6DBEF}74.51 & \cellcolor[HTML]{C6DBEF}69.61 & \cellcolor[HTML]{C6DBEF}70.59 & \cellcolor[HTML]{C6DBEF}75.49 & \cellcolor[HTML]{C6DBEF}64.71 \\
			& & Depression-P & \cellcolor[HTML]{A8D1E4}71.50 & \cellcolor[HTML]{A8D1E4}72.53 & \cellcolor[HTML]{A8D1E4}70.87 & \cellcolor[HTML]{A8D1E4}75.76 & \cellcolor[HTML]{A8D1E4}73.01 & \cellcolor[HTML]{A8D1E4}74.29 & \cellcolor[HTML]{9ECAE1}72.80 & \cellcolor[HTML]{9ECAE1}76.23 & \cellcolor[HTML]{9ECAE1}65.59 & \cellcolor[HTML]{9ECAE1}73.52 & \cellcolor[HTML]{9ECAE1}74.56 & \cellcolor[HTML]{9ECAE1}64.91 & \cellcolor[HTML]{C6DBEF}76.55 & \cellcolor[HTML]{C6DBEF}74.56 & \cellcolor[HTML]{C6DBEF}71.29 & \cellcolor[HTML]{C6DBEF}73.52 & \cellcolor[HTML]{C6DBEF}75.77 & \cellcolor[HTML]{C6DBEF}64.71 \\
			& & Depression-R & \cellcolor[HTML]{A8D1E4}70.59 & \cellcolor[HTML]{A8D1E4}70.59 & \cellcolor[HTML]{A8D1E4}72.55 & \cellcolor[HTML]{A8D1E4}73.53 & \cellcolor[HTML]{A8D1E4}72.55 & \cellcolor[HTML]{A8D1E4}73.53 & \cellcolor[HTML]{9ECAE1}72.94 & \cellcolor[HTML]{9ECAE1}74.51 & \cellcolor[HTML]{9ECAE1}66.67 & \cellcolor[HTML]{9ECAE1}70.59 & \cellcolor[HTML]{9ECAE1}74.51 & \cellcolor[HTML]{9ECAE1}64.71 & \cellcolor[HTML]{C6DBEF}71.76 & \cellcolor[HTML]{C6DBEF}74.51 & \cellcolor[HTML]{C6DBEF}66.67 & \cellcolor[HTML]{C6DBEF}70.59 & \cellcolor[HTML]{C6DBEF}75.49 & \cellcolor[HTML]{C6DBEF}64.71 \\
			& & Depression-F1      & \cellcolor[HTML]{A8D1E4}70.50 & \cellcolor[HTML]{A8D1E4}71.26 & \cellcolor[HTML]{A8D1E4}70.40 & \cellcolor[HTML]{A8D1E4}68.97 & \cellcolor[HTML]{A8D1E4}73.57 & \cellcolor[HTML]{A8D1E4}70.97 & \cellcolor[HTML]{9ECAE1}72.62 & \cellcolor[HTML]{9ECAE1}75.25 & \cellcolor[HTML]{9ECAE1}65.69 & \cellcolor[HTML]{9ECAE1}64.29 & \cellcolor[HTML]{9ECAE1}73.98 & \cellcolor[HTML]{9ECAE1}66.67 & \cellcolor[HTML]{C6DBEF}73.93 & \cellcolor[HTML]{C6DBEF}74.49 & \cellcolor[HTML]{C6DBEF}68.75 & \cellcolor[HTML]{C6DBEF}64.29 & \cellcolor[HTML]{C6DBEF}74.24 & \cellcolor[HTML]{C6DBEF}64.71 \\
			& & Control-F1         & \cellcolor[HTML]{A8D1E4}70.32 & \cellcolor[HTML]{A8D1E4}71.67 & \cellcolor[HTML]{A8D1E4}67.61 & \cellcolor[HTML]{A8D1E4}76.92 & \cellcolor[HTML]{A8D1E4}71.23 & \cellcolor[HTML]{A8D1E4}75.68 & \cellcolor[HTML]{9ECAE1}71.88 & \cellcolor[HTML]{9ECAE1}75.66 & \cellcolor[HTML]{9ECAE1}65.15 & \cellcolor[HTML]{9ECAE1}75.00 & \cellcolor[HTML]{9ECAE1}75.01 & \cellcolor[HTML]{9ECAE1}62.50 & \cellcolor[HTML]{C6DBEF}75.36 & \cellcolor[HTML]{C6DBEF}74.49 & \cellcolor[HTML]{C6DBEF}70.32 & \cellcolor[HTML]{C6DBEF}75.00 & \cellcolor[HTML]{C6DBEF}76.61 & \cellcolor[HTML]{C6DBEF}64.71 \\
			\cmidrule(lr){2-21}
			& \multirow{6}{*}{Qwen-VL} 
			& Macro-F1             & \cellcolor[HTML]{A8D1E4}79.97 & \cellcolor[HTML]{A8D1E4}83.31 & \cellcolor[HTML]{A8D1E4}76.92 & \cellcolor[HTML]{A8D1E4}85.28 & \cellcolor[HTML]{A8D1E4}84.28 & \cellcolor[HTML]{A8D1E4}85.28 & \cellcolor[HTML]{9ECAE1}79.91 & \cellcolor[HTML]{9ECAE1}86.27 & \cellcolor[HTML]{9ECAE1}80.32 & \cellcolor[HTML]{9ECAE1}88.24 & \cellcolor[HTML]{9ECAE1}83.29 & \cellcolor[HTML]{9ECAE1}85.28 & \cellcolor[HTML]{C6DBEF}78.73 & \cellcolor[HTML]{C6DBEF}84.25 & \cellcolor[HTML]{C6DBEF}74.26 & \cellcolor[HTML]{C6DBEF}67.39 & \cellcolor[HTML]{C6DBEF}81.28 & \cellcolor[HTML]{C6DBEF}85.28 \\
			& & Accuracy                & \cellcolor[HTML]{A8D1E4}80.00 & \cellcolor[HTML]{A8D1E4}83.33 & \cellcolor[HTML]{A8D1E4}77.45 & \cellcolor[HTML]{A8D1E4}85.29 & \cellcolor[HTML]{A8D1E4}84.31 & \cellcolor[HTML]{A8D1E4}85.29 & \cellcolor[HTML]{9ECAE1}80.00 & \cellcolor[HTML]{9ECAE1}86.27 & \cellcolor[HTML]{9ECAE1}80.39 & \cellcolor[HTML]{9ECAE1}88.24 & \cellcolor[HTML]{9ECAE1}83.33 & \cellcolor[HTML]{9ECAE1}85.29 & \cellcolor[HTML]{C6DBEF}78.82 & \cellcolor[HTML]{C6DBEF}84.31 & \cellcolor[HTML]{C6DBEF}74.51 & \cellcolor[HTML]{C6DBEF}67.65 & \cellcolor[HTML]{C6DBEF}81.37 & \cellcolor[HTML]{C6DBEF}85.29 \\
			& & Depression-P & \cellcolor[HTML]{A8D1E4}81.14 & \cellcolor[HTML]{A8D1E4}82.93 & \cellcolor[HTML]{A8D1E4}83.80 & \cellcolor[HTML]{A8D1E4}85.42 & \cellcolor[HTML]{A8D1E4}84.55 & \cellcolor[HTML]{A8D1E4}85.42 & \cellcolor[HTML]{9ECAE1}81.59 & \cellcolor[HTML]{9ECAE1}84.97 & \cellcolor[HTML]{9ECAE1}84.70 & \cellcolor[HTML]{9ECAE1}88.24 & \cellcolor[HTML]{9ECAE1}83.68 & \cellcolor[HTML]{9ECAE1}85.42 & \cellcolor[HTML]{C6DBEF}83.31 & \cellcolor[HTML]{C6DBEF}84.72 & \cellcolor[HTML]{C6DBEF}80.85 & \cellcolor[HTML]{C6DBEF}68.21 & \cellcolor[HTML]{C6DBEF}81.99 & \cellcolor[HTML]{C6DBEF}85.42 \\
			& & Depression-R & \cellcolor[HTML]{A8D1E4}78.82 & \cellcolor[HTML]{A8D1E4}84.31 & \cellcolor[HTML]{A8D1E4}72.55 & \cellcolor[HTML]{A8D1E4}85.29 & \cellcolor[HTML]{A8D1E4}84.31 & \cellcolor[HTML]{A8D1E4}85.29 & \cellcolor[HTML]{9ECAE1}77.65 & \cellcolor[HTML]{9ECAE1}88.24 & \cellcolor[HTML]{9ECAE1}74.51 & \cellcolor[HTML]{9ECAE1}88.24 & \cellcolor[HTML]{9ECAE1}83.33 & \cellcolor[HTML]{9ECAE1}85.29 & \cellcolor[HTML]{C6DBEF}72.94 & \cellcolor[HTML]{C6DBEF}84.31 & \cellcolor[HTML]{C6DBEF}64.71 & \cellcolor[HTML]{C6DBEF}67.65 & \cellcolor[HTML]{C6DBEF}81.37 & \cellcolor[HTML]{C6DBEF}85.29 \\
			& & Depression-F1      & \cellcolor[HTML]{A8D1E4}79.83 & \cellcolor[HTML]{A8D1E4}83.51 & \cellcolor[HTML]{A8D1E4}76.12 & \cellcolor[HTML]{A8D1E4}84.85 & \cellcolor[HTML]{A8D1E4}83.65 & \cellcolor[HTML]{A8D1E4}85.71 & \cellcolor[HTML]{9ECAE1}79.29 & \cellcolor[HTML]{9ECAE1}86.55 & \cellcolor[HTML]{9ECAE1}79.22 & \cellcolor[HTML]{9ECAE1}88.24 & \cellcolor[HTML]{9ECAE1}82.45 & \cellcolor[HTML]{9ECAE1}84.85 & \cellcolor[HTML]{C6DBEF}77.61 & \cellcolor[HTML]{C6DBEF}84.19 & \cellcolor[HTML]{C6DBEF}71.81 & \cellcolor[HTML]{C6DBEF}64.52 & \cellcolor[HTML]{C6DBEF}79.97 & \cellcolor[HTML]{C6DBEF}84.85 \\
			& & Control-F1         & \cellcolor[HTML]{A8D1E4}80.11 & \cellcolor[HTML]{A8D1E4}83.11 & \cellcolor[HTML]{A8D1E4}77.73 & \cellcolor[HTML]{A8D1E4}85.71 & \cellcolor[HTML]{A8D1E4}84.92 & \cellcolor[HTML]{A8D1E4}84.85 & \cellcolor[HTML]{9ECAE1}80.54 & \cellcolor[HTML]{9ECAE1}85.98 & \cellcolor[HTML]{9ECAE1}81.42 & \cellcolor[HTML]{9ECAE1}88.24 & \cellcolor[HTML]{9ECAE1}84.13 & \cellcolor[HTML]{9ECAE1}85.71 & \cellcolor[HTML]{C6DBEF}79.84 & \cellcolor[HTML]{C6DBEF}84.31 & \cellcolor[HTML]{C6DBEF}76.71 & \cellcolor[HTML]{C6DBEF}70.27 & \cellcolor[HTML]{C6DBEF}82.58 & \cellcolor[HTML]{C6DBEF}85.71 \\
			\midrule
			
			\multirow{12}{*}{\textbf{Transcript}} & \multirow{6}{*}{DeBERTa} 
			& Macro-F1             & \cellcolor[HTML]{F7FBFF}60.94 & \cellcolor[HTML]{F7FBFF}64.33 & \cellcolor[HTML]{F7FBFF}51.92 & \cellcolor[HTML]{F7FBFF}46.32 & \cellcolor[HTML]{F7FBFF}58.02 & \cellcolor[HTML]{F7FBFF}52.28 & \cellcolor[HTML]{E6F0F9}66.36 & \cellcolor[HTML]{E6F0F9}62.16 & \cellcolor[HTML]{E6F0F9}48.54 & \cellcolor[HTML]{E6F0F9}55.54 & \cellcolor[HTML]{E6F0F9}52.87 & \cellcolor[HTML]{E6F0F9}51.43 & \cellcolor[HTML]{EFF5FB}57.83 & \cellcolor[HTML]{EFF5FB}56.54 & \cellcolor[HTML]{EFF5FB}56.37 & \cellcolor[HTML]{EFF5FB}62.64 & \cellcolor[HTML]{EFF5FB}56.44 & \cellcolor[HTML]{EFF5FB}55.84 \\
			& & Accuracy                & \cellcolor[HTML]{F7FBFF}61.18 & \cellcolor[HTML]{F7FBFF}64.71 & \cellcolor[HTML]{F7FBFF}51.96 & \cellcolor[HTML]{F7FBFF}47.06 & \cellcolor[HTML]{F7FBFF}58.82 & \cellcolor[HTML]{F7FBFF}52.94 & \cellcolor[HTML]{E6F0F9}66.47 & \cellcolor[HTML]{E6F0F9}63.73 & \cellcolor[HTML]{E6F0F9}50.00 & \cellcolor[HTML]{E6F0F9}55.88 & \cellcolor[HTML]{E6F0F9}53.92 & \cellcolor[HTML]{E6F0F9}52.94 & \cellcolor[HTML]{EFF5FB}59.41 & \cellcolor[HTML]{EFF5FB}56.86 & \cellcolor[HTML]{EFF5FB}57.84 & \cellcolor[HTML]{EFF5FB}64.71 & \cellcolor[HTML]{EFF5FB}56.86 & \cellcolor[HTML]{EFF5FB}55.88 \\
			& & Depression-P & \cellcolor[HTML]{F7FBFF}62.90 & \cellcolor[HTML]{F7FBFF}62.32 & \cellcolor[HTML]{F7FBFF}51.85 & \cellcolor[HTML]{F7FBFF}46.89 & \cellcolor[HTML]{F7FBFF}59.55 & \cellcolor[HTML]{F7FBFF}53.11 & \cellcolor[HTML]{E6F0F9}65.76 & \cellcolor[HTML]{E6F0F9}60.08 & \cellcolor[HTML]{E6F0F9}50.21 & \cellcolor[HTML]{E6F0F9}56.07 & \cellcolor[HTML]{E6F0F9}54.30 & \cellcolor[HTML]{E6F0F9}53.36 & \cellcolor[HTML]{EFF5FB}64.86 & \cellcolor[HTML]{EFF5FB}56.85 & \cellcolor[HTML]{EFF5FB}56.39 & \cellcolor[HTML]{EFF5FB}68.89 & \cellcolor[HTML]{EFF5FB}57.16 & \cellcolor[HTML]{EFF5FB}55.90 \\
			& & Depression-R & \cellcolor[HTML]{F7FBFF}54.12 & \cellcolor[HTML]{F7FBFF}74.51 & \cellcolor[HTML]{F7FBFF}54.90 & \cellcolor[HTML]{F7FBFF}47.06 & \cellcolor[HTML]{F7FBFF}58.82 & \cellcolor[HTML]{F7FBFF}52.94 & \cellcolor[HTML]{E6F0F9}69.41 & \cellcolor[HTML]{E6F0F9}82.35 & \cellcolor[HTML]{E6F0F9}64.71 & \cellcolor[HTML]{E6F0F9}55.88 & \cellcolor[HTML]{E6F0F9}53.92 & \cellcolor[HTML]{E6F0F9}52.94 & \cellcolor[HTML]{EFF5FB}41.18 & \cellcolor[HTML]{EFF5FB}56.86 & \cellcolor[HTML]{EFF5FB}74.51 & \cellcolor[HTML]{EFF5FB}64.71 & \cellcolor[HTML]{EFF5FB}56.86 & \cellcolor[HTML]{EFF5FB}55.88 \\
			& & Depression-F1      & \cellcolor[HTML]{F7FBFF}58.11 & \cellcolor[HTML]{F7FBFF}67.84 & \cellcolor[HTML]{F7FBFF}53.33 & \cellcolor[HTML]{F7FBFF}52.63 & \cellcolor[HTML]{F7FBFF}63.79 & \cellcolor[HTML]{F7FBFF}57.89 & \cellcolor[HTML]{E6F0F9}67.39 & \cellcolor[HTML]{E6F0F9}69.27 & \cellcolor[HTML]{E6F0F9}56.37 & \cellcolor[HTML]{E6F0F9}59.46 & \cellcolor[HTML]{E6F0F9}59.81 & \cellcolor[HTML]{E6F0F9}60.00 & \cellcolor[HTML]{EFF5FB}50.00 & \cellcolor[HTML]{EFF5FB}56.44 & \cellcolor[HTML]{EFF5FB}64.07 & \cellcolor[HTML]{EFF5FB}53.85 & \cellcolor[HTML]{EFF5FB}52.19 & \cellcolor[HTML]{EFF5FB}54.55 \\
			& & Control-F1         & \cellcolor[HTML]{F7FBFF}63.76 & \cellcolor[HTML]{F7FBFF}60.83 & \cellcolor[HTML]{F7FBFF}50.51 & \cellcolor[HTML]{F7FBFF}40.00 & \cellcolor[HTML]{F7FBFF}52.26 & \cellcolor[HTML]{F7FBFF}46.67 & \cellcolor[HTML]{E6F0F9}65.33 & \cellcolor[HTML]{E6F0F9}55.06 & \cellcolor[HTML]{E6F0F9}40.72 & \cellcolor[HTML]{E6F0F9}51.61 & \cellcolor[HTML]{E6F0F9}45.93 & \cellcolor[HTML]{E6F0F9}42.86 & \cellcolor[HTML]{EFF5FB}65.66 & \cellcolor[HTML]{EFF5FB}56.65 & \cellcolor[HTML]{EFF5FB}48.67 & \cellcolor[HTML]{EFF5FB}71.43 & \cellcolor[HTML]{EFF5FB}60.69 & \cellcolor[HTML]{EFF5FB}57.14 \\
			\cmidrule(lr){2-21}
			& \multirow{6}{*}{Qwen} 
			& Macro-F1             & \cellcolor[HTML]{F7FBFF}60.57 & \cellcolor[HTML]{F7FBFF}64.69 & \cellcolor[HTML]{F7FBFF}58.16 & \cellcolor[HTML]{F7FBFF}60.92 & \cellcolor[HTML]{F7FBFF}61.37 & \cellcolor[HTML]{F7FBFF}61.73 & \cellcolor[HTML]{E6F0F9}67.69 & \cellcolor[HTML]{E6F0F9}68.38 & \cellcolor[HTML]{E6F0F9}56.81 & \cellcolor[HTML]{E6F0F9}58.88 & \cellcolor[HTML]{E6F0F9}67.58 & \cellcolor[HTML]{E6F0F9}67.39 & \cellcolor[HTML]{EFF5FB}55.07 & \cellcolor[HTML]{EFF5FB}59.29 & \cellcolor[HTML]{EFF5FB}53.83 & \cellcolor[HTML]{EFF5FB}76.14 & \cellcolor[HTML]{EFF5FB}62.13 & \cellcolor[HTML]{EFF5FB}64.21 \\
			& & Accuracy                & \cellcolor[HTML]{F7FBFF}61.18 & \cellcolor[HTML]{F7FBFF}64.71 & \cellcolor[HTML]{F7FBFF}58.82 & \cellcolor[HTML]{F7FBFF}61.76 & \cellcolor[HTML]{F7FBFF}61.76 & \cellcolor[HTML]{F7FBFF}61.76 & \cellcolor[HTML]{E6F0F9}68.24 & \cellcolor[HTML]{E6F0F9}68.63 & \cellcolor[HTML]{E6F0F9}56.86 & \cellcolor[HTML]{E6F0F9}61.76 & \cellcolor[HTML]{E6F0F9}68.63 & \cellcolor[HTML]{E6F0F9}67.65 & \cellcolor[HTML]{EFF5FB}58.24 & \cellcolor[HTML]{EFF5FB}60.78 & \cellcolor[HTML]{EFF5FB}57.84 & \cellcolor[HTML]{EFF5FB}76.47 & \cellcolor[HTML]{EFF5FB}62.75 & \cellcolor[HTML]{EFF5FB}64.71 \\
			& & Depression-P & \cellcolor[HTML]{F7FBFF}65.04 & \cellcolor[HTML]{F7FBFF}64.05 & \cellcolor[HTML]{F7FBFF}57.02 & \cellcolor[HTML]{F7FBFF}62.88 & \cellcolor[HTML]{F7FBFF}62.28 & \cellcolor[HTML]{F7FBFF}61.81 & \cellcolor[HTML]{E6F0F9}70.08 & \cellcolor[HTML]{E6F0F9}70.75 & \cellcolor[HTML]{E6F0F9}56.99 & \cellcolor[HTML]{E6F0F9}66.35 & \cellcolor[HTML]{E6F0F9}71.27 & \cellcolor[HTML]{E6F0F9}68.21 & \cellcolor[HTML]{EFF5FB}57.29 & \cellcolor[HTML]{EFF5FB}59.13 & \cellcolor[HTML]{EFF5FB}53.73 & \cellcolor[HTML]{EFF5FB}78.02 & \cellcolor[HTML]{EFF5FB}63.04 & \cellcolor[HTML]{EFF5FB}65.57 \\
			& & Depression-R & \cellcolor[HTML]{F7FBFF}52.94 & \cellcolor[HTML]{F7FBFF}66.67 & \cellcolor[HTML]{F7FBFF}70.59 & \cellcolor[HTML]{F7FBFF}61.76 & \cellcolor[HTML]{F7FBFF}61.76 & \cellcolor[HTML]{F7FBFF}61.76 & \cellcolor[HTML]{E6F0F9}63.53 & \cellcolor[HTML]{E6F0F9}64.71 & \cellcolor[HTML]{E6F0F9}54.90 & \cellcolor[HTML]{E6F0F9}61.76 & \cellcolor[HTML]{E6F0F9}68.63 & \cellcolor[HTML]{E6F0F9}67.65 & \cellcolor[HTML]{EFF5FB}57.65 & \cellcolor[HTML]{EFF5FB}74.51 & \cellcolor[HTML]{EFF5FB}54.90 & \cellcolor[HTML]{EFF5FB}76.47 & \cellcolor[HTML]{EFF5FB}62.75 & \cellcolor[HTML]{EFF5FB}64.71 \\
			& & Depression-F1      & \cellcolor[HTML]{F7FBFF}57.65 & \cellcolor[HTML]{F7FBFF}65.32 & \cellcolor[HTML]{F7FBFF}62.99 & \cellcolor[HTML]{F7FBFF}55.17 & \cellcolor[HTML]{F7FBFF}60.21 & \cellcolor[HTML]{F7FBFF}60.61 & \cellcolor[HTML]{E6F0F9}65.71 & \cellcolor[HTML]{E6F0F9}67.08 & \cellcolor[HTML]{E6F0F9}55.88 & \cellcolor[HTML]{E6F0F9}48.00 & \cellcolor[HTML]{E6F0F9}61.79 & \cellcolor[HTML]{E6F0F9}64.52 & \cellcolor[HTML]{EFF5FB}53.26 & \cellcolor[HTML]{EFF5FB}65.35 & \cellcolor[HTML]{EFF5FB}50.43 & \cellcolor[HTML]{EFF5FB}78.95 & \cellcolor[HTML]{EFF5FB}57.88 & \cellcolor[HTML]{EFF5FB}60.00 \\
			& & Control-F1         & \cellcolor[HTML]{F7FBFF}63.49 & \cellcolor[HTML]{F7FBFF}64.05 & \cellcolor[HTML]{F7FBFF}53.33 & \cellcolor[HTML]{F7FBFF}66.67 & \cellcolor[HTML]{F7FBFF}62.53 & \cellcolor[HTML]{F7FBFF}62.86 & \cellcolor[HTML]{E6F0F9}69.67 & \cellcolor[HTML]{E6F0F9}69.68 & \cellcolor[HTML]{E6F0F9}57.73 & \cellcolor[HTML]{E6F0F9}69.77 & \cellcolor[HTML]{E6F0F9}73.36 & \cellcolor[HTML]{E6F0F9}70.27 & \cellcolor[HTML]{EFF5FB}56.87 & \cellcolor[HTML]{EFF5FB}53.24 & \cellcolor[HTML]{EFF5FB}57.22 & \cellcolor[HTML]{EFF5FB}73.33 & \cellcolor[HTML]{EFF5FB}66.38 & \cellcolor[HTML]{EFF5FB}68.42 \\
			\midrule
			
			\multirow{6}{*}{\textbf{fNIRS}} & \multirow{6}{*}{Statistics} 
			& Macro-F1             & \cellcolor[HTML]{D5E4F5}62.54 & \cellcolor[HTML]{D5E4F5}71.27 & \cellcolor[HTML]{D5E4F5}59.61 & \cellcolor[HTML]{D5E4F5}43.33 & \cellcolor[HTML]{D5E4F5}73.49 & \cellcolor[HTML]{D5E4F5}61.46 & \cellcolor[HTML]{F7FBFF}65.98 & \cellcolor[HTML]{F7FBFF}69.06 & \cellcolor[HTML]{F7FBFF}54.83 & \cellcolor[HTML]{F7FBFF}42.05 & \cellcolor[HTML]{F7FBFF}73.30 & \cellcolor[HTML]{F7FBFF}45.36 & \cellcolor[HTML]{DEEBF7}62.55 & \cellcolor[HTML]{DEEBF7}64.05 & \cellcolor[HTML]{DEEBF7}65.57 & \cellcolor[HTML]{DEEBF7}50.18 & \cellcolor[HTML]{DEEBF7}75.43 & \cellcolor[HTML]{DEEBF7}46.88 \\
			& & Accuracy                & \cellcolor[HTML]{D5E4F5}62.94 & \cellcolor[HTML]{D5E4F5}71.57 & \cellcolor[HTML]{D5E4F5}60.78 & \cellcolor[HTML]{D5E4F5}52.94 & \cellcolor[HTML]{D5E4F5}73.53 & \cellcolor[HTML]{D5E4F5}61.76 & \cellcolor[HTML]{F7FBFF}67.65 & \cellcolor[HTML]{F7FBFF}69.61 & \cellcolor[HTML]{F7FBFF}56.86 & \cellcolor[HTML]{F7FBFF}47.06 & \cellcolor[HTML]{F7FBFF}73.53 & \cellcolor[HTML]{F7FBFF}47.06 & \cellcolor[HTML]{DEEBF7}62.94 & \cellcolor[HTML]{DEEBF7}64.71 & \cellcolor[HTML]{DEEBF7}66.67 & \cellcolor[HTML]{DEEBF7}52.94 & \cellcolor[HTML]{DEEBF7}75.49 & \cellcolor[HTML]{DEEBF7}47.06 \\
			& & Depression-P & \cellcolor[HTML]{D5E4F5}65.49 & \cellcolor[HTML]{D5E4F5}68.91 & \cellcolor[HTML]{D5E4F5}64.06 & \cellcolor[HTML]{D5E4F5}59.14 & \cellcolor[HTML]{D5E4F5}73.68 & \cellcolor[HTML]{D5E4F5}62.14 & \cellcolor[HTML]{F7FBFF}79.67 & \cellcolor[HTML]{F7FBFF}77.27 & \cellcolor[HTML]{F7FBFF}62.10 & \cellcolor[HTML]{F7FBFF}45.50 & \cellcolor[HTML]{F7FBFF}74.31 & \cellcolor[HTML]{F7FBFF}46.64 & \cellcolor[HTML]{DEEBF7}62.76 & \cellcolor[HTML]{DEEBF7}64.28 & \cellcolor[HTML]{DEEBF7}66.57 & \cellcolor[HTML]{DEEBF7}53.78 & \cellcolor[HTML]{DEEBF7}75.77 & \cellcolor[HTML]{DEEBF7}47.02 \\
			& & Depression-R & \cellcolor[HTML]{D5E4F5}54.12 & \cellcolor[HTML]{D5E4F5}78.43 & \cellcolor[HTML]{D5E4F5}60.78 & \cellcolor[HTML]{D5E4F5}52.94 & \cellcolor[HTML]{D5E4F5}73.53 & \cellcolor[HTML]{D5E4F5}61.76 & \cellcolor[HTML]{F7FBFF}47.06 & \cellcolor[HTML]{F7FBFF}56.86 & \cellcolor[HTML]{F7FBFF}37.25 & \cellcolor[HTML]{F7FBFF}47.06 & \cellcolor[HTML]{F7FBFF}73.53 & \cellcolor[HTML]{F7FBFF}47.06 & \cellcolor[HTML]{DEEBF7}63.53 & \cellcolor[HTML]{DEEBF7}68.63 & \cellcolor[HTML]{DEEBF7}72.55 & \cellcolor[HTML]{DEEBF7}52.94 & \cellcolor[HTML]{DEEBF7}75.49 & \cellcolor[HTML]{DEEBF7}47.06 \\
			& & Depression-F1      & \cellcolor[HTML]{D5E4F5}59.07 & \cellcolor[HTML]{D5E4F5}72.96 & \cellcolor[HTML]{D5E4F5}59.99 & \cellcolor[HTML]{D5E4F5}20.00 & \cellcolor[HTML]{D5E4F5}72.77 & \cellcolor[HTML]{D5E4F5}58.06 & \cellcolor[HTML]{F7FBFF}58.70 & \cellcolor[HTML]{F7FBFF}65.25 & \cellcolor[HTML]{F7FBFF}45.92 & \cellcolor[HTML]{F7FBFF}25.00 & \cellcolor[HTML]{F7FBFF}70.88 & \cellcolor[HTML]{F7FBFF}35.71 & \cellcolor[HTML]{DEEBF7}62.56 & \cellcolor[HTML]{DEEBF7}65.34 & \cellcolor[HTML]{DEEBF7}67.72 & \cellcolor[HTML]{DEEBF7}38.46 & \cellcolor[HTML]{DEEBF7}74.24 & \cellcolor[HTML]{DEEBF7}43.75 \\
			& & Control-F1         & \cellcolor[HTML]{D5E4F5}66.00 & \cellcolor[HTML]{D5E4F5}69.58 & \cellcolor[HTML]{D5E4F5}59.23 & \cellcolor[HTML]{D5E4F5}66.67 & \cellcolor[HTML]{D5E4F5}74.22 & \cellcolor[HTML]{D5E4F5}64.86 & \cellcolor[HTML]{F7FBFF}73.25 & \cellcolor[HTML]{F7FBFF}72.86 & \cellcolor[HTML]{F7FBFF}63.75 & \cellcolor[HTML]{F7FBFF}59.09 & \cellcolor[HTML]{F7FBFF}75.71 & \cellcolor[HTML]{F7FBFF}55.00 & \cellcolor[HTML]{DEEBF7}62.54 & \cellcolor[HTML]{DEEBF7}62.76 & \cellcolor[HTML]{DEEBF7}63.43 & \cellcolor[HTML]{DEEBF7}61.90 & \cellcolor[HTML]{DEEBF7}76.61 & \cellcolor[HTML]{DEEBF7}50.00 \\
			\bottomrule
		\end{tabular}
	}
	\caption{Performance of different models and feature sets, evaluated for each task and modality combination.	}
	\label{tab:performance_metrics}
\end{table*}


\begin{table*}[htbp]
	\centering
	
	
	\definecolor{gColorC1}{RGB}{178, 213, 235} 
	\definecolor{gColorC2}{RGB}{221, 237, 248} 
	\definecolor{gColorC3}{RGB}{175, 211, 235} 
	\definecolor{gColorC4}{RGB}{241, 248, 253} 
	\definecolor{gColorC5}{RGB}{162, 205, 231} 
	\definecolor{gColorC6}{RGB}{216, 234, 246} 
	\definecolor{gColorC7}{RGB}{219, 236, 247} 
	\definecolor{gColorC8}{RGB}{157, 202, 229} 
	\definecolor{gColorC9}{RGB}{247, 251, 255} 
	\definecolor{gColorC10}{RGB}{162, 205, 231} 
	\definecolor{gColorC11}{RGB}{153, 200, 228} 
	\definecolor{gColorC12}{RGB}{153, 200, 228} 
	\definecolor{gColorC13}{RGB}{192, 221, 239} 
	\definecolor{gColorC14}{RGB}{144, 195, 225} 
	\definecolor{gColorC15}{RGB}{142, 194, 224} 
	
	\definecolor{gColorF1}{RGB}{153, 200, 228} 
	\definecolor{gColorF2}{RGB}{244, 249, 254} 
	\definecolor{gColorF3}{RGB}{133, 189, 222} 
	\definecolor{gColorF4}{RGB}{240, 247, 253} 
	\definecolor{gColorF5}{RGB}{229, 241, 250} 
	\definecolor{gColorF6}{RGB}{133, 189, 222} 
	\definecolor{gColorF7}{RGB}{202, 226, 242} 
	\definecolor{gColorF8}{RGB}{122, 184, 219} 
	\definecolor{gColorF9}{RGB}{182, 215, 237} 
	\definecolor{gColorF10}{RGB}{143, 194, 225} 
	\definecolor{gColorF11}{RGB}{107, 174, 214} 
	\definecolor{gColorF12}{RGB}{195, 223, 240} 
	\definecolor{gColorF13}{RGB}{122, 184, 219} 
	\definecolor{gColorF14}{RGB}{113, 179, 216} 
	\definecolor{gColorF15}{RGB}{112, 178, 216} 
	
	\renewcommand{\arraystretch}{1.6}
	\setlength{\tabcolsep}{2.4mm}
	\resizebox{\textwidth}{!}{%
		\begin{tabular}{@{}llrcccccccccccccc!{\vrule width 0pt}}
			\toprule
			\textbf{Feature Set} & \textbf{Metric} & \textbf{A} & \textbf{T} & \textbf{V} & \textbf{N} & \textbf{\begin{tabular}{@{}c@{}}A\\V\end{tabular}} & \textbf{\begin{tabular}{@{}c@{}}A\\N\end{tabular}} & \textbf{\begin{tabular}{@{}c@{}}T\\A\end{tabular}} & \textbf{\begin{tabular}{@{}c@{}}T\\V\end{tabular}} & \textbf{\begin{tabular}{@{}c@{}}T\\N\end{tabular}} & \textbf{\begin{tabular}{@{}c@{}}V\\N\end{tabular}} & \textbf{\begin{tabular}{@{}c@{}}A\\V\\N\end{tabular}} & \textbf{\begin{tabular}{@{}c@{}}T\\A\\V\end{tabular}} & \textbf{\begin{tabular}{@{}c@{}}T\\A\\N\end{tabular}} & \textbf{\begin{tabular}{@{}c@{}}T\\V\\N\end{tabular}} & \textbf{\begin{tabular}{@{}c@{}}T\\A\\V\\N\end{tabular}} \\
			\midrule
			\multirow{2}{*}{Classific} & Mean & \cellcolor{gColorC1}72.98 & \cellcolor{gColorC2}68.24 & \cellcolor{gColorC3}73.24 & \cellcolor{gColorC4}66.33 & \cellcolor{gColorC5}74.47 & \cellcolor{gColorC6}68.76 & \cellcolor{gColorC7}68.51 & \cellcolor{gColorC8}75.00 & \cellcolor{gColorC9}65.72 & \cellcolor{gColorC10}74.46 & \cellcolor{gColorC11}75.39 & \cellcolor{gColorC12}75.39 & \cellcolor{gColorC13}71.53 & \cellcolor{gColorC14}76.22 & \cellcolor{gColorC15}76.47 \\
			& Std & \cellcolor{gColorC1}$\pm$6.60 & \cellcolor{gColorC2}$\pm$1.71 & \cellcolor{gColorC3}$\pm$6.01 & \cellcolor{gColorC4}$\pm$1.84 & \cellcolor{gColorC5}$\pm$1.66 & \cellcolor{gColorC6}$\pm$1.78 & \cellcolor{gColorC7}$\pm$8.91 & \cellcolor{gColorC8}$\pm$7.61 & \cellcolor{gColorC9}$\pm$8.07 & \cellcolor{gColorC10}$\pm$4.53 & \cellcolor{gColorC11}$\pm$6.93 & \cellcolor{gColorC12}$\pm$6.93 & \cellcolor{gColorC13}$\pm$4.52 & \cellcolor{gColorC14}$\pm$0.44 & \cellcolor{gColorC15}$\pm$0.00 \\
			\midrule
			\multirow{2}{*}{Foundation} & Mean & \cellcolor{gColorF1}75.43 & \cellcolor{gColorF2}66.05 & \cellcolor{gColorF3}77.30 & \cellcolor{gColorF4}66.38 & \cellcolor{gColorF5}67.53 & \cellcolor{gColorF6}77.27 & \cellcolor{gColorF7}70.43 & \cellcolor{gColorF8}78.39 & \cellcolor{gColorF9}72.50 & \cellcolor{gColorF10}76.38 & \cellcolor{gColorF11}80.36 & \cellcolor{gColorF12}71.10 & \cellcolor{gColorF13}78.42 & \cellcolor{gColorF14}79.30 & \cellcolor{gColorF15}79.39 \\
			& Std & \cellcolor{gColorF1}$\pm$6.13 & \cellcolor{gColorF2}$\pm$6.84 & \cellcolor{gColorF3}$\pm$7.47 & \cellcolor{gColorF4}$\pm$4.40 & \cellcolor{gColorF5}$\pm$0.24 & \cellcolor{gColorF6}$\pm$1.46 & \cellcolor{gColorF7}$\pm$2.95 & \cellcolor{gColorF8}$\pm$1.74 & \cellcolor{gColorF9}$\pm$6.14 & \cellcolor{gColorF10}$\pm$2.87 & \cellcolor{gColorF11}$\pm$1.67 & \cellcolor{gColorF12}$\pm$1.60 & \cellcolor{gColorF13}$\pm$1.69 & \cellcolor{gColorF14}$\pm$0.08 & \cellcolor{gColorF15}$\pm$0.00 \\
			\bottomrule
		\end{tabular}%
	}
	\caption{Performance comparison of modality fusion.
		\label{tab:model_performance_final}
	}
\end{table*}

\begin{table*}[htbp]
	\centering
	
	\definecolor{s2ColorC1}{RGB}{183, 216, 236} \definecolor{s2ColorC2}{RGB}{247, 251, 255} \definecolor{s2ColorC3}{RGB}{224, 238, 248} \definecolor{s2ColorC4}{RGB}{142, 194, 224} \definecolor{s2ColorC5}{RGB}{196, 223, 240} \definecolor{s2ColorC6}{RGB}{172, 210, 234} \definecolor{s2ColorC7}{RGB}{141, 193, 224} \definecolor{s2ColorF1}{RGB}{137, 191, 223} \definecolor{s2ColorF2}{RGB}{123, 185, 219} \definecolor{s2ColorF3}{RGB}{133, 189, 222} \definecolor{s2ColorF4}{RGB}{122, 184, 219} \definecolor{s2ColorF5}{RGB}{131, 188, 221} \definecolor{s2ColorF6}{RGB}{130, 188, 221} \definecolor{s2ColorF7}{RGB}{107, 174, 214}
	
	\renewcommand{\arraystretch}{1.2}
	\setlength{\tabcolsep}{7.7mm}
	\resizebox{\textwidth}{!}{%
		\begin{tabular}{@{}llccccccc!{\vrule width 0pt}}
			\toprule
			\textbf{Feature Set} & \textbf{Metric} & \textbf{INT} & \textbf{PDT} & \textbf{VFT} & \textbf{\begin{tabular}{@{}c@{}}INT\\PDT\end{tabular}} & \textbf{\begin{tabular}{@{}c@{}}INT\\VFT\end{tabular}} & \textbf{\begin{tabular}{@{}c@{}}PDT\\VFT\end{tabular}} & \textbf{\begin{tabular}{@{}c@{}}INT\\PDT\\VFT\end{tabular}} \\
			\midrule
			\multirow{2}{*}{Classific} & Mean & \cellcolor{s2ColorC1}73.49 & \cellcolor{s2ColorC2}68.51 & \cellcolor{s2ColorC3}70.31 & \cellcolor{s2ColorC4}76.37 & \cellcolor{s2ColorC5}72.43 & \cellcolor{s2ColorC6}74.32 & \cellcolor{s2ColorC7}76.46 \\
			& Std & \cellcolor{s2ColorC1}$\pm$2.90 & \cellcolor{s2ColorC2}$\pm$8.34 & \cellcolor{s2ColorC3}$\pm$3.29 & \cellcolor{s2ColorC4}$\pm$3.04 & \cellcolor{s2ColorC5}$\pm$4.59 & \cellcolor{s2ColorC6}$\pm$1.58 & \cellcolor{s2ColorC7}$\pm$2.94 \\
			\midrule
			\multirow{2}{*}{Foundation Model} & Mean & \cellcolor{s2ColorF1}76.78 & \cellcolor{s2ColorF2}78.23 & \cellcolor{s2ColorF3}77.17 & \cellcolor{s2ColorF4}78.34 & \cellcolor{s2ColorF5}77.38 & \cellcolor{s2ColorF6}77.42 & \cellcolor{s2ColorF7}79.38 \\
			& Std & \cellcolor{s2ColorF1}$\pm$4.90 & \cellcolor{s2ColorF2}$\pm$4.81 & \cellcolor{s2ColorF3}$\pm$4.62 & \cellcolor{s2ColorF4}$\pm$1.70 & \cellcolor{s2ColorF5}$\pm$4.58 & \cellcolor{s2ColorF6}$\pm$4.51 & \cellcolor{s2ColorF7}$\pm$2.98 \\
			\bottomrule
		\end{tabular}%
	}
	\caption{Performance comparison of task fusion.
		\label{tab:model_performance_set2_fixed}
	}
\end{table*}

\begin{table*}[!htbp]
	\centering
	
	\definecolor{mincolor}{HTML}{F7FBFF}
	\definecolor{maxcolor}{HTML}{6BAED6}
	
	
	
	\colorlet{cDZ1}{maxcolor!59.67!mincolor}  
	\colorlet{cDZ2}{maxcolor!55.44!mincolor}  
	\colorlet{cDZ3}{maxcolor!60.97!mincolor}  
	\colorlet{cDZ4}{maxcolor!54.56!mincolor}  
	\colorlet{cDZ5}{maxcolor!58.56!mincolor}  
	\colorlet{cDZ6}{maxcolor!68.56!mincolor}  
	\colorlet{cDZ7}{maxcolor!85.39!mincolor}  
	\colorlet{cDF1}{maxcolor!54.9!mincolor}   
	\colorlet{cDF2}{maxcolor!84.32!mincolor}  
	\colorlet{cDF3}{maxcolor!50.48!mincolor}  
	\colorlet{cDF4}{maxcolor!51.85!mincolor}  
	\colorlet{cDF5}{maxcolor!85.85!mincolor}  
	\colorlet{cDF6}{maxcolor!80.43!mincolor}  
	\colorlet{cDF7}{maxcolor!2.9!mincolor}    
	
	\colorlet{cVZ1}{maxcolor!82.6!mincolor}   
	\colorlet{cVZ2}{maxcolor!43.57!mincolor}  
	\colorlet{cVZ3}{maxcolor!65.66!mincolor}  
	\colorlet{cVZ4}{maxcolor!55.6!mincolor}   
	\colorlet{cVZ5}{maxcolor!62.15!mincolor}  
	\colorlet{cVZ6}{maxcolor!73.79!mincolor}  
	\colorlet{cVZ7}{maxcolor!28.69!mincolor}  
	\colorlet{cVF1}{maxcolor!82.79!mincolor}  
	\colorlet{cVF2}{maxcolor!48.61!mincolor}  
	\colorlet{cVF3}{maxcolor!62.46!mincolor}  
	\colorlet{cVF4}{maxcolor!100.0!mincolor}  
	\colorlet{cVF5}{maxcolor!0.0!mincolor}    
	\colorlet{cVF6}{maxcolor!86.99!mincolor}  
	\colorlet{cVF7}{maxcolor!22.28!mincolor}  
	
	\colorlet{cPZ1}{maxcolor!86.41!mincolor}  
	\colorlet{cPZ2}{maxcolor!64.14!mincolor}  
	\colorlet{cPZ3}{maxcolor!74.97!mincolor}  
	\colorlet{cPZ4}{maxcolor!67.91!mincolor}  
	\colorlet{cPZ5}{maxcolor!67.45!mincolor}  
	\colorlet{cPZ6}{maxcolor!90.5!mincolor}   
	\colorlet{cPZ7}{maxcolor!32.35!mincolor}  
	\colorlet{cPF1}{maxcolor!86.71!mincolor}  
	\colorlet{cPF2}{maxcolor!70.96!mincolor}  
	\colorlet{cPF3}{maxcolor!68.83!mincolor}  
	\colorlet{cPF4}{maxcolor!88.44!mincolor}  
	\colorlet{cPF5}{maxcolor!40.21!mincolor}  
	\colorlet{cPF6}{maxcolor!89.81!mincolor}  
	\colorlet{cPF7}{maxcolor!0.23!mincolor}   
	
	\newcommand{\rot}[1]{\rotatebox{60}{#1}}
	\renewcommand{\arraystretch}{1.4}
	
	\begin{adjustbox}{width=\textwidth}
		\begin{tabular}{@{}llcccccccccccccc!{\vrule width 0pt}}
			\toprule
			\multicolumn{1}{c}{\textbf{Strategy}} & \multicolumn{1}{c}{\textbf{Metric}} & \multicolumn{7}{c}{\textbf{Zero-Shot}} & \multicolumn{7}{c}{\textbf{Few-shot}} \\
			\cmidrule(lr){3-9} \cmidrule(lr){10-16}
			& & \rot{GPT-4o} & \rot{DeepSeek-v3} & \rot{Qwen3(NT)} & \rot{GPT-o3} & \rot{DeepSeek-r1} & \rot{Qwen3(T)} & \rot{Qwen2.5-Omni} & \rot{GPT-4o} & \rot{DeepSeek-v3} & \rot{Qwen3(NT)} & \rot{GPT-o3} & \rot{DeepSeek-r1} & \rot{Qwen3(T)} & \rot{Qwen2.5-Omni} \\
			\midrule
			
			\multirow{6}{*}{\begin{tabular}[c]{@{}c@{}}Direct \\ Prediction\end{tabular}} 
			& Macro-F1             & \cellcolor{cDZ1}53.52 & \cellcolor{cDZ2}52.41 & \cellcolor{cDZ3}53.87 & \cellcolor{cDZ4}52.18 & \cellcolor{cDZ5}53.23 & \cellcolor{cDZ6}55.85 & \cellcolor{cDZ7}60.26 & \cellcolor{cDF1}52.27 & \cellcolor{cDF2}59.98 & \cellcolor{cDF3}51.11 & \cellcolor{cDF4}51.47 & \cellcolor{cDF5}60.38 & \cellcolor{cDF6}58.95 & \cellcolor{cDF7}38.64 \\
			& Accuracy                  & \cellcolor{cDZ1}53.54 & \cellcolor{cDZ2}52.53 & \cellcolor{cDZ3}54.55 & \cellcolor{cDZ4}52.53 & \cellcolor{cDZ5}53.54 & \cellcolor{cDZ6}56.57 & \cellcolor{cDZ7}62.63 & \cellcolor{cDF1}57.58 & \cellcolor{cDF2}60.61 & \cellcolor{cDF3}51.52 & \cellcolor{cDF4}51.52 & \cellcolor{cDF5}61.62 & \cellcolor{cDF6}60.61 & \cellcolor{cDF7}45.45 \\
			& Depression-P & \cellcolor{cDZ1}63.37 & \cellcolor{cDZ2}63.1  & \cellcolor{cDZ3}70    & \cellcolor{cDZ4}65    & \cellcolor{cDZ5}61.34 & \cellcolor{cDZ6}62.94 & \cellcolor{cDZ7}65.17 & \cellcolor{cDF1}60    & \cellcolor{cDF2}66.8  & \cellcolor{cDF3}63.64 & \cellcolor{cDF4}60    & \cellcolor{cDF5}66.25 & \cellcolor{cDF6}64.44 & \cellcolor{cDF7}52.00    \\
			& Depression-R    & \cellcolor{cDZ1}45.61 & \cellcolor{cDZ2}42.11 & \cellcolor{cDZ3}36.84 & \cellcolor{cDZ4}38.6  & \cellcolor{cDZ5}52.63 & \cellcolor{cDZ6}59.65 & \cellcolor{cDZ7}75.44 & \cellcolor{cDF1}78.95 & \cellcolor{cDF2}63.16 & \cellcolor{cDF3}36.84 & \cellcolor{cDF4}47.37 & \cellcolor{cDF5}68.42 & \cellcolor{cDF6}70.18 & \cellcolor{cDF7}68.42 \\
			& Depression-F1        & \cellcolor{cDZ1}53.03 & \cellcolor{cDZ2}50.44 & \cellcolor{cDZ3}48.28 & \cellcolor{cDZ4}48.35 & \cellcolor{cDZ5}56.57 & \cellcolor{cDZ6}61.15 & \cellcolor{cDZ7}69.91 & \cellcolor{cDF1}68.18 & \cellcolor{cDF2}64.9  & \cellcolor{cDF3}46.67 & \cellcolor{cDF4}52.94 & \cellcolor{cDF5}67.28 & \cellcolor{cDF6}67.18 & \cellcolor{cDF7}59.09 \\
			& Control-F1           & \cellcolor{cDZ1}54.01 & \cellcolor{cDZ2}54.37 & \cellcolor{cDZ3}59.46 & \cellcolor{cDZ4}56.01 & \cellcolor{cDZ5}49.9  & \cellcolor{cDZ6}50.54 & \cellcolor{cDZ7}50.62 & \cellcolor{cDF1}36.36 & \cellcolor{cDF2}55.06 & \cellcolor{cDF3}55.56 & \cellcolor{cDF4}50    & \cellcolor{cDF5}53.48 & \cellcolor{cDF6}50.71 & \cellcolor{cDF7}18.18 \\
			\midrule
			
			\multirow{6}{*}{\begin{tabular}[c]{@{}c@{}}Vanilla \\ Reasoning\end{tabular}}
			& Macro-F1             & \cellcolor{cVZ1}59.53 & \cellcolor{cVZ2}49.3  & \cellcolor{cVZ3}55.09 & \cellcolor{cVZ4}52.45 & \cellcolor{cVZ5}54.17 & \cellcolor{cVZ6}57.22 & \cellcolor{cVZ7}45.4  & \cellcolor{cVF1}59.58 & \cellcolor{cVF2}50.62 & \cellcolor{cVF3}54.25 & \cellcolor{cVF4}64.09 & \cellcolor{cVF5}37.88 & \cellcolor{cVF6}60.69 & \cellcolor{cVF7}43.72 \\
			& Accuracy                  & \cellcolor{cVZ1}59.6  & \cellcolor{cVZ2}49.49 & \cellcolor{cVZ3}55.56 & \cellcolor{cVZ4}52.53 & \cellcolor{cVZ5}54.55 & \cellcolor{cVZ6}57.58 & \cellcolor{cVZ7}45.45 & \cellcolor{cVF1}59.6  & \cellcolor{cVF2}51.52 & \cellcolor{cVF3}54.55 & \cellcolor{cVF4}64.65 & \cellcolor{cVF5}56.57 & \cellcolor{cVF6}61.62 & \cellcolor{cVF7}45.45 \\
			& Depression-P & \cellcolor{cVZ1}71.89 & \cellcolor{cVZ2}59.01 & \cellcolor{cVZ3}62.96 & \cellcolor{cVZ4}63.46 & \cellcolor{cVZ5}62.2  & \cellcolor{cVZ6}64.71 & \cellcolor{cVZ7}53.16 & \cellcolor{cVF1}70.7  & \cellcolor{cVF2}65.66 & \cellcolor{cVF3}62.58 & \cellcolor{cVF4}70.41 & \cellcolor{cVF5}57.27 & \cellcolor{cVF6}66.7  & \cellcolor{cVF7}52.46 \\
			& Depression-R    & \cellcolor{cVZ1}49.12 & \cellcolor{cVZ2}42.11 & \cellcolor{cVZ3}56.14 & \cellcolor{cVZ4}42.11 & \cellcolor{cVZ5}54.39 & \cellcolor{cVZ6}57.89 & \cellcolor{cVZ7}40.35 & \cellcolor{cVF1}50.88 & \cellcolor{cVF2}33.33 & \cellcolor{cVF3}50.88 & \cellcolor{cVF4}66.67 & \cellcolor{cVF5}96.49 & \cellcolor{cVF6}66.67 & \cellcolor{cVF7}54.39 \\
			& Depression-F1        & \cellcolor{cVZ1}58.31 & \cellcolor{cVZ2}48.86 & \cellcolor{cVZ3}59.25 & \cellcolor{cVZ4}50.6  & \cellcolor{cVZ5}57.95 & \cellcolor{cVZ6}61.11 & \cellcolor{cVZ7}45.83 & \cellcolor{cVF1}59.15 & \cellcolor{cVF2}44.13 & \cellcolor{cVF3}55.79 & \cellcolor{cVF4}68.45 & \cellcolor{cVF5}71.85 & \cellcolor{cVF6}66.65 & \cellcolor{cVF7}53.35 \\
			& Control-F1           & \cellcolor{cVZ1}60.76 & \cellcolor{cVZ2}49.73 & \cellcolor{cVZ3}50.93 & \cellcolor{cVZ4}54.29 & \cellcolor{cVZ5}50.4  & \cellcolor{cVZ6}53.33 & \cellcolor{cVZ7}44.98 & \cellcolor{cVF1}60.01 & \cellcolor{cVF2}57.12 & \cellcolor{cVF3}52.7  & \cellcolor{cVF4}59.74 & \cellcolor{cVF5}3.92  & \cellcolor{cVF6}54.72 & \cellcolor{cVF7}34.1  \\
			\midrule
			
			\multirow{6}{*}{\begin{tabular}[c]{@{}c@{}}Psychiatric \\ Reasoning\end{tabular}} 
			& Macro-F1             & \cellcolor{cPZ1}60.53 & \cellcolor{cPZ2}54.69 & \cellcolor{cPZ3}57.54 & \cellcolor{cPZ4}55.68 & \cellcolor{cPZ5}55.56 & \cellcolor{cPZ6}61.6  & \cellcolor{cPZ7}46.36 & \cellcolor{cPF1}60.61 & \cellcolor{cPF2}56.48 & \cellcolor{cPF3}55.91 & \cellcolor{cPF4}61.06 & \cellcolor{cPF5}48.42 & \cellcolor{cPF6}61.42 & \cellcolor{cPF7}37.94 \\
			& Accuracy                  & \cellcolor{cPZ1}60.61 & \cellcolor{cPZ2}55.56 & \cellcolor{cPZ3}57.58 & \cellcolor{cPZ4}56.57 & \cellcolor{cPZ5}64.55 & \cellcolor{cPZ6}61.62 & \cellcolor{cPZ7}46.46 & \cellcolor{cPF1}60.61 & \cellcolor{cPF2}56.57 & \cellcolor{cPF3}57.58 & \cellcolor{cPF4}61.62 & \cellcolor{cPF5}58.59 & \cellcolor{cPF6}61.62 & \cellcolor{cPF7}38.38 \\
			& Depression-P & \cellcolor{cPZ1}72.52 & \cellcolor{cPZ2}72.58 & \cellcolor{cPZ3}69.23 & \cellcolor{cPZ4}75.19 & \cellcolor{cPZ5}52.63 & \cellcolor{cPZ6}73.08 & \cellcolor{cPZ7}54.96 & \cellcolor{cPF1}71.43 & \cellcolor{cPF2}65.79 & \cellcolor{cPF3}62.26 & \cellcolor{cPF4}67.94 & \cellcolor{cPF5}59.28 & \cellcolor{cPF6}69.36 & \cellcolor{cPF7}44.44 \\
			& Depression-R    & \cellcolor{cPZ1}50.88 & \cellcolor{cPZ2}36.84 & \cellcolor{cPZ3}47.37 & \cellcolor{cPZ4}36.84 & \cellcolor{cPZ5}57.83 & \cellcolor{cPZ6}52.63 & \cellcolor{cPZ7}42.11 & \cellcolor{cPF1}52.63 & \cellcolor{cPF2}50.88 & \cellcolor{cPF3}66.67 & \cellcolor{cPF4}63.16 & \cellcolor{cPF5}89.47 & \cellcolor{cPF6}59.65 & \cellcolor{cPF7}26.32 \\
			& Depression-F1        & \cellcolor{cPZ1}59.67 & \cellcolor{cPZ2}48.68 & \cellcolor{cPZ3}56.25 & \cellcolor{cPZ4}49.43 & \cellcolor{cPZ5}52.75 & \cellcolor{cPZ6}61.17 & \cellcolor{cPZ7}47.59 & \cellcolor{cPF1}60.61 & \cellcolor{cPF2}57.28 & \cellcolor{cPF3}64.33 & \cellcolor{cPF4}65.31 & \cellcolor{cPF5}71.31 & \cellcolor{cPF6}64.13 & \cellcolor{cPF7}33.00    \\
			& Control-F1           & \cellcolor{cPZ1}61.39 & \cellcolor{cPZ2}60.7  & \cellcolor{cPZ3}58.82 & \cellcolor{cPZ4}61.93 & \cellcolor{cPZ5}55.29 & \cellcolor{cPZ6}62.03 & \cellcolor{cPZ7}45.13 & \cellcolor{cPF1}60.61 & \cellcolor{cPF2}55.68 & \cellcolor{cPF3}47.48 & \cellcolor{cPF4}56.81 & \cellcolor{cPF5}25.54 & \cellcolor{cPF6}58.71 & \cellcolor{cPF7}42.88 \\
			\bottomrule
		\end{tabular}
	\end{adjustbox}
	
	\caption{Performance of different LLMs and reasoning strategies.}
	\label{tab:strategy_metric_comparison_colored}
	
\end{table*}

\section{Experimental Results Details}

This section provides supplementary tables with detailed results for the experiments. The detailed performance metrics for the behavioral signature modeling, including Precision, Recall, and F1-scores, are presented in Table~\ref{tab:performance_metrics}. The detailed results for the modality and task fusion are shown in Table~\ref{tab:model_performance_final} and Table~\ref{tab:model_performance_set2_fixed}. The results for the psychiatric reasoning experiments with LLMs, which show the performance of different models under various reasoning strategies, are available in Table~\ref{tab:strategy_metric_comparison_colored}.

\end{document}